\documentclass[twoside,11pt]{article}

\usepackage{blindtext}

\usepackage[preprint]{jmlr2e}

\usepackage[margin=1.5in]{geometry}
\usepackage{natbib}
\setcitestyle{round}

\usepackage[utf8]{inputenc} %
\usepackage[T1]{fontenc}    %
\usepackage{lmodern}
\usepackage{hyperref}       %

\usepackage{inconsolata}
\usepackage{url}            %
\usepackage{booktabs}       %
\usepackage{amsfonts}       %
\usepackage{nicefrac}       %
\usepackage{microtype}      %
\usepackage{marginalia}

\usepackage{array,amssymb,latexsym,amssymb,epsfig,epsf,amsmath,cite,accents,float,mathrsfs,verbatim,bm,color,enumitem,bbm}
\graphicspath{{figures/}}
\usepackage[linesnumbered,noend]{algorithm2e}
\usepackage{appendix}
\SetKwInput{KwEn}{Encoding}
\SetKwInput{KwDe}{Decoding}

\usepackage{wrapfig}
\usepackage{transparent}

\usepackage[font=small,labelfont=bf]{caption}
\UseRawInputEncoding

\newtheorem{assumption}{Assumption}

\newcommand{\ud}{\,\mathrm{d}}
\newcommand{\erf}{\,\mathrm{erf}}

\def\ubf{{\bf u}}
\def\vbf{{\bf v}}
\def\wbf{{\bf w}}

\def\ybf{{\bf y}}
\def\zbf{{\bf z}}

\def\ybf{{\bf y}}

\def\Pbf{{\bf P}}

\def\Jc{{\cal J}}

\def\Oc{{\cal O}}

\def\Sc{{\cal S}}
\def\Tc{{\cal T}}

\def\Zc{{\cal Z}}

\def\zsbf{{\boldsymbol{\mathsf z}}}
\def\wsbf{{\boldsymbol{\mathsf w}}}
\def\psbf{{\boldsymbol{\mathsf p}}}
\def\qsbf{{\boldsymbol{\mathsf q}}}
\def\ssbf{{\boldsymbol{\mathsf s}}}
\def\xsbf{{\boldsymbol{\mathsf x}}}
\def\ysbf{{\boldsymbol{\mathsf y}}}

\def\isf{{\mathsf i}}
\def\rsf{{\mathsf r}}
\def\wsf{{\mathsf w}}

\def\nn{\nonumber}
\def\beq{\begin{equation}}
\def\eeq{\end{equation}}
\def\beqa{\begin{eqnarray}}
\def\eeqa{\end{eqnarray}}
\def\balign{\begin{align}}
\def\ealign{\end{align}}
\def\bpr{\begin{proof}}
\def\epr{\end{proof}}
\def\bth{\begin{theorem}}
\def\eth{\end{theorem}}
\def\blm{\begin{lemma}}
\def\elm{\end{lemma}}
\def\bprop{\begin{proposition}}
\def\eprop{\end{proposition}}
\def\bcr{\begin{corollary}}
\def\ecr{\end{corollary}}

\def\ie{{\it i.e.,\ \/}}
\def\eg{{\it e.g.,\ \/}}
\def\defeq{:=}

\def\opt{{\rm opt}}

\def\E{\mathbb{E}}

\def\and {{\rm and}}

\def\loss{\ell}

\def\ind{\mathbbm{1}}

\usepackage{xr-hyper}
\usepackage{xcolor}
\hypersetup{hidelinks}

\numberwithin{equation}{section}		%
\usepackage[sort&compress,capitalize,nameinlink]{cleveref}
\crefname{assumption}{Assumption}{Assumptions}
\crefname{lemma}{Lemma}{Lemmas}

\usepackage{lastpage}
\jmlrheading{25}{2024}{1-\pageref{LastPage}}{1/22; Revised 10/23}{1/24}{22-0068}{Ali Ramezani-Kebrya, Kimon Antonakopoulos, Volkan Cevher, Ashish Khisti, and Ben Liang}

\ShortHeadings{On the Generalization of SGD with Momentum}{Ramezani-Kebrya, Antonakopoulos, Cevher, Khisti, and Liang}
\firstpageno{1}

\begin{document}

\title{On the Generalization of Stochastic\\ Gradient Descent with Momentum}

\author{Ali Ramezani-Kebrya\email ali@uio.no \\
     \addr Department of Informatics, University of Oslo and Visual Intelligence Centre\\
     Integreat, Norwegian Centre for Knowledge-driven Machine Learning\\
     Gaustadall\'{e}en 23B, Ole-Johan Dahls hus, 0373 Oslo, Norway
     \AND
      \name Kimon Antonakopoulos\email kimon.antonakopoulos@epfl.ch\\
        \addr Laboratory for Information and Inference Systems (LIONS), EPFL\\
       EPFL STI IEL LIONS, Station 11, CH-1015 Lausanne, Switzerland
        \AND
        \name Volkan Cevher\email volkan.cevher@epfl.ch\\
        \addr Laboratory for Information and Inference Systems (LIONS), EPFL\\
       EPFL STI IEL LIONS, Station 11, CH-1015 Lausanne, Switzerland
        \AND        
       \name Ashish Khisti \email akhisti@ece.utoronto.ca \\
       \addr Department of Electrical and Computer
        Engineering, University of Toronto\\
       40 St. George Street, Toronto, ON M5S 2E4, Canada
       \AND
       \name Ben Liang \email liang@ece.utoronto.ca \\
       \addr Department of Electrical and Computer
        Engineering, University of Toronto\\
       40 St. George Street, Toronto, ON M5S 2E4, Canada
       }

\maketitle

\begin{abstract}%
While momentum-based {accelerated variants of} %
stochastic gradient descent (SGD) are widely used when training machine learning models, there is little theoretical understanding on the generalization error of such methods. In this work, we first show that there exists a convex loss function for which the stability gap for multiple epochs of SGD with standard heavy-ball momentum (SGDM) becomes unbounded. Then, for smooth Lipschitz loss functions, we analyze a modified momentum-based update rule, \ie SGD with early momentum (SGDEM) {under a broad range of step-sizes}, and show that {it can train machine learning models for multiple epochs with a guarantee for generalization.}
Finally, for the special case of strongly convex loss functions, we find a range of momentum such that multiple epochs of standard SGDM, as a special form of SGDEM, also generalizes. Extending our results on generalization, we also develop an upper bound on the expected true risk, in terms of the number of training steps, sample size, and momentum.  Our  experimental evaluations verify the consistency between the numerical results and our theoretical bounds. SGDEM improves the generalization error of SGDM  when training ResNet-18 on ImageNet in practical distributed settings.\end{abstract}

\begin{keywords}
  Uniform stability, generalization error,  heavy-ball momentum, stochastic gradient descent,  non-convex 
\end{keywords}

\section{Introduction}\label{sec:intro}

    \begin{wrapfigure}{R}{0.4\textwidth}
        \vspace*{-0.7cm}
            \includegraphics[width=0.39\textwidth]{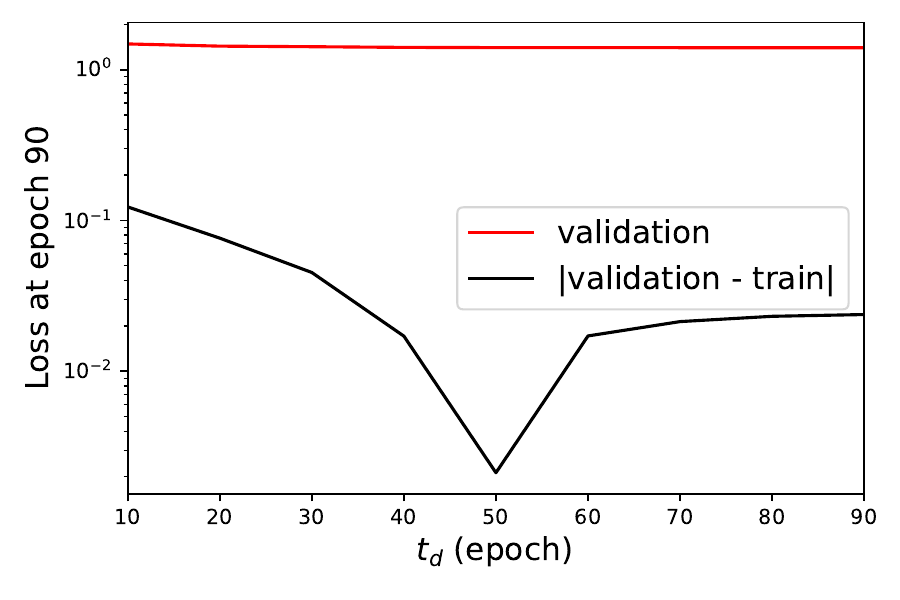}
            \caption{Validation loss and generalization error of SGDEM when training ResNet-18~ \citep{Resnet} on ImageNet~\citep{ImageNet}  in a distributed setting with 4 GPUs under tuned step-size and global minibatch size of 128. For each $t_d$, the momentum is set to  $\mu_d=0.9$ in the first $t_d$ epochs and then zero for the next $90-t_d$ epochs.  SGDM is a special form of SGDEM with $t_d=90$.  The details are provided in~\cref{sec:numerical} and~\cref{app:exp} .}
            \label{fig:ImageNet_testgenerrvst}
        \vspace*{-0.4cm}
    \end{wrapfigure}

Stochastic gradient descent (SGD) and its variants are the most popular algorithms for training deep neural networks due to their support of efficient parallel implementations and excellent generalization performance~\citep{Krizhevsky,wilson2017marginal}. %
To accelerate the convergence of SGD, a momentum term is often added in the iterative update of the stochastic gradient \citep{DLbook}. This approach has a long history, with proven benefits in various settings. The heavy-ball momentum method was first introduced by \citet{Polyak} where a weighted version of the previous update is added to the current gradient update. \citet{Polyak} motivated his method by its resemblance to a heavy ball moving in a potential well defined by the objective function. Momentum methods have been used  to accelerate empirical risk minimization when training neural networks \citep{Rumelhart}. In particular, momentum methods are used for training deep neural networks with complex and nonconvex loss landscapes~\citep{Sutskever}. Intuitively, adding momentum accelerates convergence by circumventing sharp curvatures and long ravines of the sub-level sets of the objective function \citep{Wilson}.  \citet{Ochs15} present an illustrative example  to show that the momentum can potentially avoid local minima.

Beyond convergence, the generalization of machine learning algorithms is a fundamental problem in learning theory. A classical framework used to study the generalization error in machine learning is PAC learning \citep{VC,PAC}. However, the associated bounds using this approach can be conservative. The connection between stability and generalization has been studied in the literature, which captures how the learning algorithm explores a hypothesis class~\citep{Bousquet,SSSS,Hardt}. According to the definition of \citet{Bousquet}, uniform stability requires the algorithm to generate almost the same predictions for all datasets that are different in only one example. Recently, this notion of uniform stability is leveraged to analyze the generalization error of SGD \citep{Hardt}. \citet{Hardt} have derived the stability bounds for SGD and analyzed its generalization for different loss functions. This is a substantial step forward, since SGD is widely used in many practical systems.
However, the algorithms studied in these  works do not include momentum.

In this work, we study SGD with momentum (SGDM). Although {momentum methods are empirically observed to accelerate training in deep learning \citep{DLbook}}, their effect on the generalization error is not well understood. Even though momentum is not studied  in \citep{Hardt}, it is conjectured therein  that momentum might speed up training but adversely impact generalization.

By providing a counter example, we show that the stability gap for multiple epochs of SGDM can become unbounded even for convex loss functions. This motivates us to consider a modified momentum-based update rule, called SGD with \emph{early momentum} (SGDEM) where a momentum term is added in the earlier training steps. We show that SGDEM is guaranteed to generalize for {smooth Lipschitz loss functions} and any momentum. 
To the best of our knowledge, stability and generalization of SGDEM have not been considered in the existing literature.  As~\cref{fig:ImageNet_testgenerrvst} shows in a practical and distributed setting on ImageNet, while validation loss remains unaffected, the minimum generalization error happens if the momentum is applied for 50 epochs, which indicates that tuning momentum is useful to achieve the best generalization error. 

We study the generalization error and true risk of SGDEM. In order to find an upper bound on the expected generalization error of SGDEM, we use the framework of uniform stability \citep{Bousquet,Hardt}. %

\subsection{Main Contributions}
In~\cref{sec:earlym}, we show that there exists a convex loss function for which the stability gap for multiple epochs of SGDM becomes unbounded. We introduce SGDEM and show that it is guaranteed to generalize for {smooth Lipschitz loss} functions.
 We obtain a bound on the generalization error of SGDEM that decreases inversely with the size of the training set. Our results show that the number of iterations can grow as $n^l$ for a small $l>1$ where $n$ is the sample size, which explains why complicated models such as deep neural networks can be trained for multiple epochs of SGDEM while their generalization errors are limited.~{We also establish an explicit convergence rate for SGDEM and smooth Lipschitz
loss functions under a broad range of hyperparameters including a general step-size rule that covers popular step-sizes in the optimization literature. Our convergence and generalization bounds capture the inherent trade-off between optimization and generalization.}

In~\cref{sec:sconvex}, we focus on the special case of strongly convex loss functions. In this case, we show that one can obtain a bound on the generalization error of standard SGDM, which suggests that this special form of SGDEM suffices for generalization. Our bound is independent of the number of training iterations and decreases inversely with the size of the training set. Finally, we establish an upper bound on the expected true risk of SGDM as a function of various problem parameters. %

Our generalization bounds for both strongly convex and {smooth Lipschitz loss} functions tend to zero as the number of samples increases. In addition, our results confirm that using a momentum parameter, $\mu\approx 1$, for the entire training improves optimization error { under certain settings}. However, it adversely affects the generalization error bounds. Hence, it is crucial to establish an appropriate balance between the optimization error associated with the empirical risk and the generalization error. 

Finally, our experimental results show that SGDEM outperforms both vanilla SGD and SGDM in terms of test error on CIFAR10 and generalization error on ImageNet. %

\subsection{Related Work}

Studies on the generalization of momentum methods are scarce in the literature. As explained above, momentum is not considered in \citep{Bousquet,Hardt}. While the generalization error of SGDM is studied in \citep{Ong} and \citep{Chen}, their analysis is limited to the special case of quadratic loss functions. In this work, we show that unlike SGDM, multiple epochs of SGDEM is guaranteed to generalize for {smooth Lipschitz loss} functions. A similar hybrid method has been shown to generalize better than both vanilla SGD and Adaptive Moment Estimation (Adam)  in deep learning practice \citep{SWATS}.  However, it remains unclear why such hybrid methods generalize better. Our work sheds theoretical light on this question.

Convergence of first-order methods with momentum has been studied in~\citep{Polyak,Ochs14,Ochs15,Ghadimi,Lessard,Yang,Wilson,Gadat,Orvieto,Can}. Most of these works consider the deterministic setting for gradient update \citep{Polyak,Ochs14,Ochs15,Ghadimi,Lessard,Wilson}. %
Only a few works have analyzed convergence in the stochastic setting \citep{Yang,Gadat,Orvieto,Can}. In \citep{Yang}, a unified convergence analysis of SGDM has been studied for both convex and nonconvex loss functions with bounded variance. \citet{Gadat} have studied the almost sure convergence results of the stochastic heavy-ball method with nonconvex coercive loss functions and provided a complexity analysis for the case of quadratic strongly convex. In \citep{Orvieto}, differential equation-based analysis is used to study convergence of SGDM. \citet{Can} have obtained linear convergence rates for SGDM under a particular momentum for the special case of quadratic loss functions. In this paper, we introduce \emph{early momentum} for the class of {smooth Lipschitz loss} functions, which requires unique convergence analysis as shown in~\cref{sec:earlym}.

We further note that \citet{Lessard} have provided a specific loss function for which the heavy-ball method does not converge. This loss function does not contradict our convergence analysis. The loss function in~\citep{Lessard} has been carefully constructed and does not satisfy the assumptions considered in this paper.

In addition to Polyak's heavy-ball momentum method, \citet{Nesterov} has proposed an accelerated gradient descent, which converges as $O(1/k^2)$ in a deterministic and convex setting where $k$ is the number of iterations. Convergence rates are obtained for Nesterov's accelerated gradient method in various settings~\citep{Su,Laborde,Assran}. However, the Netstrov momentum does not seem to improve the rate of convergence for  stochastic gradient settings~\citep[Section 8.3.3]{DLbook}. Therefore, in this work we focus on the heavy-ball momentum.

High-probability bounds on the generalization error of uniformly stable algorithms over the random choice of the dataset have been recently established in \citep{feldman2018generalization,feldman2019high,bousquet2020sharper,klochkov2021stability}. Momentum-based methods have not been considered in these works.  In this paper, we establish generalization errors for  momentum-based methods with a focus on the randomness of the algorithm.

{ Our results complement the recent results of~\citet{attia2021instability}, which show exponential growth in uniform stability bounds of accelerated gradient descent methods. We focus on stochastic gradient descent with heavy-ball momentum for possibly nonconvex problems under nonconstant step-sizes, while~\citet{attia2021instability} focus on convex problems with full-batch Nesterov's accelerated gradient under a fixed step-size.}

\textbf{Notation:} We use $\E[\cdot]$ to denote the expectation and $\|\cdot\|$ to represent the Euclidean norm of a vector. We use lower-case bold font to denote vectors. We use sans-serif font to denote random quantities. Sets and scalars are represented by calligraphic and standard fonts, respectively.

\section{Problem and Assumptions}\label{sec:prelim}
 We consider a general supervised learning problem, where $\Sc=\{\zsbf_1,\ldots,\zsbf_n\}$ denotes the set of samples of size $n$ drawn i.i.d. from some space $\Zc$ with an unknown distribution $D$. We assume a learning model described by parameter vector $\wbf\in \Omega$. Let $\loss(\wbf;\zbf)$ denote the loss of the model described by parameter $\wbf$ on example $\zbf\in \Zc$.

The ultimate goal of learning is to minimize the true or population risk given by
\begin{align}\label{Truerisk}
R(\wbf)\defeq \E_{\zsbf\sim D}[\loss(\wbf;\zsbf)].
\end{align} Since the distribution $D$ is unknown, we approximate this objective by the empirical risk during training, $\ie R_\Sc(\wbf)\defeq \frac{1}{n}\sum_{i=1}^n\loss(\wbf;\zsbf_i)$.
 We assume $\wsbf=A(\Sc)$ for some potentially randomized algorithm $A$.

\subsection{Generalization Error and Stability}
In order to find an upper bound on the true risk of algorithm $A$, in this work, we consider the generalization error, which is the expected difference of empirical and true risk:
\begin{align}\label{generror}
\epsilon_g\defeq \E_{\Sc,A}[R(A(\Sc))-R_\Sc(A(\Sc))].
\end{align}
In order to find an upper bound on the generalization error of algorithm $A$, we consider the uniform stability property.
\begin{definition} Let $\Sc$ and $\Sc'$ denote two datasets from space $\Zc^n$ such that $\Sc$ and $\Sc'$ differ in at most one example. Algorithm $A$ is $\epsilon_s$-uniformly stable if for all datasets $\Sc$ and $\Sc'$, we have
\begin{align}\label{stabdef}
\sup_\zbf\E_A[\loss(A(\Sc);\zbf)-\loss(A(\Sc');\zbf)]\leq\epsilon_s.
\end{align}
\end{definition}
It is known that uniform stability implies generalization in expectation:
\bth[\citealt{Hardt}]\label{genstabth} If $A$ is an $\epsilon_s$-uniformly stable algorithm, then the generalization error of $A$ is upper bounded by $\epsilon_s$.
\eth \cref{genstabth} suggests that it is enough to control the uniform stability of an algorithm to bound the generalization error.

\subsection{{SGDM}}%
The update rule for SGDM is given by
\begin{align}\label{update}\tag{SGDM}
{\wsbf}_{t+1}&=\wsbf_t+\mu(\wsbf_t-\wsbf_{t-1})-\alpha_t\nabla_\wbf \loss(\wsbf_t;\zbf_{\isf_t})
\end{align} where  $\alpha_t>0$ is the step-size,  {$\mu\in(0,1]$} is the momentum parameter, $\isf_t\in\{1,\ldots,n\}$ is a  selected index drawn uniformly at random at each iteration, and $\loss(\wsbf_t;\zbf_{\isf_t})$ is the loss evaluated on sample $\zbf_{\isf_t}$.\footnote{Another variant  to select $\isf_t$ is to permutate $\{1,\ldots,n\}$ randomly once and then select the examples repeatedly in a cyclic manner.  Our stability analysis in~\cref{sec:earlymnonconv,sec:sconvex} holds under both variants, i.e., uniformly at random with replacement and random permutation.} In~\ref{update}, we run the update iteratively for $T$ steps and let $\wsbf_T$ denote the final output.

In the case where the parameter space $\Omega$ is a compact and convex set, we consider the update rule for projected SGDM:
\begin{align}\label{projupdate}\tag{P-SGDM}
{\wsbf}_{t+1}&=\Pbf\Big(\wsbf_t+\mu(\wsbf_t-\wsbf_{t-1})-\alpha_t\nabla_\wbf \loss(\wsbf_t;\zbf_{\isf_t})\Big)
\end{align} where $\Pbf$ denotes the Euclidean projection onto $\Omega$.
The key quantity of interest in this paper is the generalization error given by
\begin{align}\nn
\epsilon_g= \E_{\Sc,A}[R(\wsbf_T)-R_\Sc(\wsbf_T)] = \E_{\Sc, \isf_0,\ldots, \isf_{T-1}}[R(\wsbf_T)-R_\Sc(\wsbf_T)]
\end{align} since the randomness in $A$ arises from the choice of $\isf_0,\ldots, \isf_{T-1}$.

\subsection{Assumptions on  Loss Function}\label{sec:assumption}
Let $\zbf\in\Zc$. In our analysis, we will assume that the loss function $\loss(\cdot;\zbf)$ satisfies the following properties, which are used also in \citep{Hardt}.

{\begin{assumption}[Lipschitzness  \& smoothness]\label{assu:LipSm}
Let $\zbf\in\Zc$. The  loss function $\loss(\cdot;\zbf)$ satisfies the following properties: 1) $L$-Lipschitzness: There exists some $L>0$ such that   $|\loss(\ubf;\zbf)-\loss(\vbf;\zbf)|\leq L\|\ubf-\vbf\|$ for all $\ubf,\vbf\in\Omega$; 2) $\beta$-smoothness: There exists some $\beta>0$ such that 
$\|\nabla \loss(\ubf;\zbf)-\nabla \loss(\vbf;\zbf)\|\leq \beta\|\ubf-\vbf\|$ for all $\ubf,\vbf\in\Omega$.
\end{assumption}}

Our assumptions hold for neural networks with  smooth activation functions such as smooth approximations of ReLU including softplus or Gaussian error Linear Units (GeLU) \citep{Dugas2000,hendrycks2016gaussian}.  We note that  softplus and GeLU typically match and exceed performance compared to ReLU \citep{clevert2015fast,Xu2015}.

\section{Smooth Lipschitz Loss}\label{sec:earlym}

We first show that there exists a convex loss function for which the stability gap for multiple epochs of~\ref{update} becomes unbounded. For the case of {smooth Lipschitz loss} functions, we introduce SGDEM and show that machine learning models can be trained for multiple epochs of SGDEM while their generalization errors are bounded. %

In SGDEM, the momentum $\mu$ is set to some constant {$ \mu_d\in(0,1]$} in the first $t_d$ steps and then zero for $t=t_d+1,\ldots,T$. Thus, the update rule for SGDEM is given by
\begin{align}\label{earlyupdate}\tag{SGDEM}
{\wsbf}_{t+1}&=\wsbf_t+\mu_d \mathbbm{1}(t\leq t_d)(\wsbf_t-\wsbf_{t-1})-\alpha_t\nabla_\wbf \loss(\wsbf_t;\zbf_{\isf_t})
\end{align} where $\mathbbm{1}$ denotes the indicator function, and the projected version can be similarly {defined} based on~\ref{projupdate}.
\vspace{-0.2cm}
\subsection{Uniform Stability Bounds for~\ref{update} and~\ref{earlyupdate}}\label{sec:lowerbound}
\begin{example}\label{ex1}
Let $w\in[-1,1]$ denote a parameter. Consider the one-dimensional and convex loss function $\loss(w;\zbf)=L_{\zbf} w +c_{\zbf}$ where $ L_{\zbf}\in\{L,-L\} $ depending on $\zbf\in\Zc$ and $c_{\zbf}\geq 0$ is constant w.r.t. $w$. For a specific choice of $\Sc$ with 
\begin{align}\nn
R_{\Sc}(w)=\frac{1}{n}\sum_{i=1}^n(Lw+c_i)=Lw+\sum_{i=1}^n \frac{c_i}{n},
\end{align}%
the optimal parameter minimizing the empirical risk is $w^*_{\Sc}=-1$.
\end{example} Both~\ref{update} and~\ref{earlyupdate} can find the optimal solution of our convex empirical risk minimization problem. We first establish a lower bound on the stability gap when~\ref{update} is run for multiple epochs, which shows that the gap can be unbounded.\footnote{If we set $T=n^l$ for some $l>1$, the stability gap will become unbounded as $n\rightarrow \infty$ regardless of $\mu$. If we set $T=k n$ with $k>1$, there will still be a nonvanishing gap regardless of $\mu$ (the gap does not blow up but it does not vanish).} {In our analysis of the stability, our goal is to track the divergence of two different iterative sequences of update rules with the same starting point.}
\bth\label{thm:lowerbound} 
Let $\zbf\in\Zc$. Suppose that the~\ref{update} update  is executed for $T$ steps with constant step-size $\alpha>0$ and  $\mu\in(0,1]$ on~\cref{ex1}.
There exist datasets $\Sc$ and $\Sc'$ such that $\E_A[\loss(A(\Sc);\zbf)-\loss(A(\Sc');\zbf)]$
is lower bounded by $\Omega(\frac{T}{n}(1+k\mu^k))$ with $k=\lceil\log(T)\rceil$.\footnote{ We note that the lower bound in~\cref{thm:lowerbound} matches an upper bound for SGD with smooth and convex losses, and early-stopped~\ref{update} will be stable  by setting $T$ sublinearly in $n$.} 
\eth

\bpr We consider two neighbouring datasets $\Sc$ and $\Sc'$ with 
\begin{align}\nn
R_{\Sc}(w)=\frac{1}{n}\sum_{i=1}^n(Lw+c_i)=Lw+\overline{c}
\end{align}
 and 
\begin{align}\nn
R_{\Sc'}(w)=-\frac{1}{n}Lw+\frac{c'_k}{n}+\frac{1}{n}\sum_{i=1,i\neq k}^n(Lw+c_i)=\frac{n-2}{n}Lw+\frac{c'_k-c_k}{n}+\overline{c} 
\end{align} where $\overline{c}=\frac{1}{n}\sum_{i=1}^n c_i$.
 Suppose we select an index $\isf_t$ uniformly at random from $\{1,\ldots,n\}$. Then we have $\E_{\isf_t}[\nabla \loss(w;\zbf_{\isf_t})]=\nabla R_{\Sc}(w)=L$ and $\E_{\isf_t}[\nabla \loss(w;\zbf'_{\isf_t})]=\nabla R_{\Sc'}(w)=(n-2)L/n$, which holds for all $w\in\Omega$. Let $\wsf_T$ and $\wsf'_T$ denote the outputs of~\ref{update} on $\Sc$ and $\Sc'$, respectively. Suppose $\wsf_0=\wsf'_0$. We can follow the steps of~\ref{update} on $\Sc$ and obtain\footnote{We assume $\wsf_0,\alpha,\mu$ are set such that the updated parameter remains in the parameter space.}
\begin{align}\nn
\E_{\isf_0,\ldots,\isf_{T-1}}[\wsf_T]=-(T+(T-1)\mu+(T-2)\mu^2+\cdots+\mu^{T-1})\alpha L+\E[\wsf_0]. 
\end{align} 
 
 Similarly, we have
\begin{align}\nn
\E_{\isf_0,\ldots,\isf_{T-1}}[\wsf'_T]=-(T+(T-1)\mu+(T-2)\mu^2+\cdots+\mu^{T-1})\frac{(n-2)\alpha L}{n}+\E[\wsf_0].
\end{align} 
 
Hence, we have 
\begin{equation}\nn
\E_A[\wsf'_T-\wsf_T]=\frac{2\alpha L}{n}(T+(T-1)\mu+(T-2)\mu^2+\cdots+\mu^{T-1}).
\end{equation} 
 
Let $\zbf\in\Zc$. Using Jensen's inequality, we can show that 
\begin{align}
\E_A[|\loss(\wsf_T;\zbf)-\loss(\wsf'_T;\zbf)|]&\geq |\E_A[\loss(\wsf_T;\zbf)-\loss(\wsf'_T;\zbf)]|\nn\\
&=\frac{2\alpha L^2}{n} \sum_{j=0}^{T-1}(T-j)\mu^j.\nn
\end{align}
Hence, $\E_A[|\loss(\wsf_T;\zbf)-\loss(\wsf'_T;\zbf)|]$ is lower bounded by $\Omega(T/n(1+k\mu^k))$ where $k=o(T)$.
\epr

Note that the stability lower bound increases monotonically with $\mu$. For the same example, we can show that the stability gap for multiple epochs of~\ref{earlyupdate}  goes to zero.

\bth\label{thm:exampleSGDEM} 
Let $\mu_d \in (0, 1)$. For~\cref{ex1} and datasets described in~\cref{thm:lowerbound}, the stability gap for~\ref{earlyupdate} and~\ref{update} goes to zero as $n\to \infty$ as long as $\sum_{j=1}^T\alpha_j=o(n)$.\\ 

\noindent Furthermore, under $T=n^l$ with some $l>1$ and $t_d\big(\sum_{j=1}^T\alpha_j^2\big)^{1/3}=o(T^{\frac{2}{3l}})$, the stability gap for~\ref{earlyupdate} goes to zero for any $\mu_d\in(0,1]$. 
\eth

\bpr  Let $\wsf_T$ and $\wsf'_T$ denote the outputs of~\ref{earlyupdate} on $\Sc$ and $\Sc'$, respectively.
Following the steps of~\ref{earlyupdate} on $\Sc$, we have 
\begin{align}\nn
\E_{\isf_0,\ldots,\isf_{T-1}}[\wsf_T]&=-L(\alpha_1+\cdots+\alpha_T)-\alpha_1L(\mu_d+\cdots+\mu_d^{t_d-1})-\alpha_2L^2(\mu_d+\cdots+\mu_d^{t_d-2})\\
&\quad-\cdots-\alpha_{t_d-1}L\mu_d+\E[\wsf_0].\nn 
\end{align}
Similarly, we have 
\begin{align}\nn
E_{\isf_0,\ldots,\isf_{T-1}}[\wsf'_T]&=-((n-2)/n)L(\alpha_1+\cdots+\alpha_T)-((n-2)/n)\alpha_1L(\mu_d+\cdots+\mu_d^{t_d-1})\\&\quad-((n-2)/n)\alpha_2L^2(\mu_d+\cdots+\mu_d^{t_d-2})-\cdots-((n-2)/n)\alpha_{t_d-1}L\mu_d+\E[\wsf_0].\nn
\end{align}
 Let $\zbf\in\Zc$. For this particular example, we have 
$\E_A[|\loss(\wsf_T;\zbf)-\loss(\wsf'_T;\zbf)|]=L\E_A|\wsf_T-\wsf'_T|.$

\blm $\wsf'_T-\wsf_T\geq 0$ everywhere.
\elm 
\bpr Let $w_T$ and $w_T'$ denote a realization of $\wsf_T$ and $\wsf'_T$, respectively. Let $w_0$ denote a realization of $\wsf_0$, and fix $(\isf_0,\ldots,\isf_{T-1})$ where the differing index between $\Sc$ and $\Sc'$ happens at steps in $\Jc=\{j_1,\ldots,j_m\}$. Then we have 

\begin{align}\nn
w_T&=-L(\alpha_1+\cdots+\alpha_T)-\alpha_1L(\mu_d+\cdots+\mu_d^{t_d-1})-\alpha_2L^2(\mu_d+\cdots+\mu_d^{t_d-2})\\&\quad-\cdots-\alpha_{t_d-1}L\mu_d+w_0\nn
\end{align}
 and 
\begin{align}\nn
w_T'&=-L\sum_{j\in\Tc/\Jc}\alpha_j+L\sum_{j\in\Jc}\alpha_j-L\sum_{j\in\Tc_d/\Jc_d}\alpha_j(\mu_d+\cdots+\mu_d^{t_d-j})\\&\quad+L\sum_{j\in\Jc_d}\alpha_j(\mu_d+\cdots+\mu_d^{t_d-j})+w_0\nn
\end{align} 
  where $\Tc=\{1,\ldots,T\}$, $\Tc_d=\{1,\ldots,t_d\}$, and $\Jc_d=\{j\in\Jc:~j\leq t_d\}$.  This shows $w_T'\geq w_T$. We note that the set of realizations with $w_T'< w_T$ is empty.
\epr
This lemma shows $\E_A|\wsf_T-\wsf'_T|=\E_A[\wsf'_T]-\E_A[\wsf_T]$. We first note that 

\begin{align}\nn
\begin{split}
\E_A[|\loss(\wsf_T;\zbf)-\loss(\wsf'_T;\zbf)|]&=\frac{2L^2}{n}\sum_{j=1}^T\alpha_j+\frac{2L^2}{n}\sum_{j=1}^{t_d}\alpha_j(\mu_d+\cdots+\mu_d^{t_d-j})\\
&\leq\frac{2L^2}{n}\sum_{j=1}^T\alpha_j+\frac{2L^2\mu_d}{n(1-\mu_d)}\sum_{j=1}^{t_d}\alpha_j\\
&\leq \frac{2L^2 \sum_{j=1}^T\alpha_j}{n}\Big(1+\frac{\mu_d}{1-\mu_d}\Big).
\end{split}
\end{align}

Finally, we have 
\begin{align}\nn
\E_A[|\loss(\wsf_T;\zbf)-\loss(\wsf'_T;\zbf)|]&=\frac{2L^2}{n}\sum_{j=1}^T\alpha_j+\frac{2L^2}{n}\sum_{j=1}^{t_d}\alpha_j(\mu_d+\cdots+\mu_d^{t_d-j})\\
&{\leq\frac{2L^2}{n}\sum_{j=1}^T\alpha_j+\frac{2L^2}{n} \sum_{j=1}^{t_d}\alpha_j(t_d-j)}\nn\\
&{\leq \frac{2L^2}{n}\sum_{j=1}^T\alpha_j+\frac{2L^2}{n} \sqrt{\frac{(t_d^2-1)(2t_d-1)}{6}\sum_{j=1}^{t_d}\alpha_j^2}}\nn
\end{align} where the last inequality holds due to Cauchy-Schwarz.  This completes the proof. 
\epr

For constant step-size, we can show an $\Omega(\frac{T}{n})$ lower bound on the stability gap for~\ref{update} even when we set momentum to zero.  However, this does not explain what happens if we use another step-size. To highlight the importance of early momentum on bounding the stability gap, in~\cref{app:thm:exampleSGDM_lr}, we show that the stability gap for multiple epochs of~\ref{update} may become unbounded for any step-size schedule. This includes $\alpha_1=1$ and  $\alpha_j=0$ for $j>1$, \ie  the gradient term is added only in the first iteration. {We also establish an $\Omega(\frac{T}{n})$ lower bound for~\ref{update} on~\cref{ex1} even with a \textit{time-decaying} step-size, {which shows that it is important to control both step-size and momentum to establish uniform stability.\footnote{Time-decaying step-size is required to establish uniform stability for multiple epochs of SGD without momentum in the convex case \citep[Theorem 3.8]{Hardt}.} In~\cref{sec:sconvex}, we show that SGDM with fixed step-size is stable for strongly convex loss functions.This demonstrates the role of the loss function.}}
 
\bcr For~\cref{ex1} with datasets described in~\cref{thm:lowerbound} and for \textit{time-decaying step-size}, the stability gap for~\ref{update} is lower bounded by $\Omega(\frac{T}{n})$ for \textit{any momentum} $\mu>0$.
\ecr
\bpr It follows immediately from the proof of~\cref{thm:exampleSGDEM}. 
\epr 

\begin{remark} 

As shown in~\cref{sec:earlymnonconv}, unlike~\ref{update},~\ref{earlyupdate} is stable and thus guaranteed to generalize for {smooth Lipschitz loss} functions and any momentum.  We remark that, since uniform stability is only a sufficient condition for generalization, our result here does not necessarily imply that~\ref{update} does not generalize.  Our results highlight that step-size schedule, momentum, and the structure of the loss play roles in establishing uniform stability. In~\cref{sec:numerical}, we show an empirical example that~\ref{update} does not generalize in a nonconvex problem.  

\end{remark}

\subsection{Generalization Analysis of~\ref{earlyupdate}}\label{sec:earlymnonconv}
Since generalization is predicated on the convergence of a learning algorithm, we first show that~\ref{earlyupdate} is guaranteed to converge to a local minimum for general and possibly nonconvex problems. Then, we show that~\ref{earlyupdate} is guaranteed to generalize for any $\mu_d$, when $t_d$ is chosen appropriately. Our analysis captures the inherent trade-off between optimization and generalization.

{ Let $q\in [\frac{1}{2},1)$. We  establish an explicit convergence rate for~\ref{earlyupdate} and a general step-size $\alpha_t=\alpha_0/t^q$, which includes as special cases popular choices of step-sizes in the optimization literature. Together with the generalization bounds, our analyses characterize the optimization error in terms of the expected norm of gradients of empirical risk and generalization error
for stochastic gradient descent with heavy-ball momentum under a broad range of hyperparameters and smooth Lipschitz
loss functions. To the best of our knowledge, this is the first work providing such results.

\bth\label{thm:generallrconv}
Let $q\in [\frac{1}{2},1)$. Suppose that  $\loss$ satisfies~\cref{assu:LipSm} and that the~\ref{earlyupdate} update is executed for $T$ steps with step-size $\alpha_t=\alpha_0/t^q$ and any $1 \leq t_d \leq T$. Then 
we have 
\begin{align}\label{generallrconvRate}
\begin{split}
\min_{1\leq t\leq T}\E_{A}\big[\|\nabla R_\Sc(\wsbf_{t})\|^2\big]
&\leq \frac{2\big(R_\Sc(\wsbf_{0}) - \inf_{\wbf} R_\Sc(\wbf)\big)}{\sum_{t=1}^{T}\alpha_t}\\
&\quad + \frac{
\max\Big\{\frac{\beta L^2}{2(1-\mu_d)^2},\frac{\beta\mu_d L^2}{(1-\mu_d)^3},\frac{\beta L^2}{2} 
\Big\}\sum_{t=1}^{T} \alpha_t^2}{\sum_{t=1}^{T}\alpha_t}.
\end{split}
\end{align}
In particular,~\ref{earlyupdate} achieves the rate of  $\Oc(T^{q-1})$ for any $t_d$.
\eth
\bpr See~\cref{app:thm:generallrconv}. 
\epr

}

{\begin{remark} Convergence of~\ref{earlyupdate} with constant step-size and another time-dependent step-size are provided in~\cref{app:thm:earlymnconconv} and~\cref{app:timelr_decay}, respectively. In~\cref{app:cor:SGDEMvstd}, we provide a sufficient condition for the optimization bound to become a monotonically decreasing function of $t_d$.  In~\cref{app:momentumrole}, we study the convergence bound for a special form of~\ref{earlyupdate} and show the benefit of using momentum. We also provide a simple sufficient condition for the non-vanishing term in the convergence bound to become a monotonically decreasing function of $\mu_d$.
\end{remark}}

We first show that for { $\alpha_t=\alpha_0/t$ and} carefully designed $t_d$,~\ref{earlyupdate} updates satisfy uniform stability, and the number of stochastic gradient steps can grow as $n^l$ for a small $l>1$ while the generalization error is limited. We note that~\ref{earlyupdate} is guaranteed to generalize for any $\mu_d$. { Then we establish an upper bound on the generalization error of~\ref{earlyupdate} for a general step-size $\alpha_t=\alpha_0/t^q$ with $q\in [\frac{1}{2},1)$.}

\bth\label{thm:earlymnconstab}
Suppose {that  $\loss$ satisfies~\cref{assu:LipSm} and} that  the~\ref{earlyupdate} update is executed for $T$ steps with step-size $\alpha_t=\alpha_0/t$ and some constant {$\mu_d\in(0,1]$} in the first $t_d$ steps. Then, for any $1 \leq \tilde t \leq t_d \leq T$,~\ref{earlyupdate} satisfies $\epsilon_s$-uniform stability with
\begin{align}\label{earlymstabnonc}
\!\!\!\!\!\!\epsilon_s\leq\frac{2\alpha_0L^2}{n}T^{u}\tilde h(\mu_d,t_d)+\frac{\tilde t M}{n}+\frac{2L^2}{\beta(n-1)}\Big(\frac{T}{\tilde t}\Big)^{u}
\end{align}
where {$\tilde h(\mu_d,t_d)=\exp(2\mu_dt_d)\big(E_1(2\mu_d\tilde t)-E_1(2\mu_d t_d)\big)$,  $E_1(x)\defeq\int_{x}^{\infty}\frac{\exp(-t)}{t}\ud t$},  $u=(1-\frac{1}{n})\alpha_0\beta$, and $M=\sup_{\wbf,\zbf}\loss(\wbf;\zbf)$. 
\eth

\bpr  Let $\Sc$ and $\Sc'$ be two sets of samples of size $n$ that differ in at most one example. Let $\wsbf_T$ and $\wsbf'_T$ denote the outputs of~\ref{update} on $\Sc$ and $\Sc'$, respectively. We consider the updates $\wsbf_{t+1}=G_t(\wsbf_t)+\mu_t(\wsbf_t-\wsbf_{t-1})$ and $\wsbf'_{t+1}=G'_t(\wsbf'_t)+\mu_t(\wsbf'_t-\wsbf'_{t-1})$ where $\mu_t\defeq\mu_d \mathbbm{1}(t\leq t_d)$ with $G_t(\wsbf_t)=\wsbf_t-\alpha_t\nabla_{\wbf}\loss(\wsbf_t;\zbf_{\isf_t})$ and $G'_t(\wsbf'_t)=\wsbf'_t-\alpha_t\nabla_{\wbf}\loss(\wsbf'_t;\zbf'_{\isf_t})$, respectively, for $t=1,\ldots,T$. We denote $\delta_t\defeq \|\wsbf_t-\wsbf'_t\|$. Suppose $\wsbf_0=\wsbf'_0$, \ie $\delta_0=0$.

First, as a preliminary step, we observe that the expected loss difference under $\wsbf_T$ and $\wsbf'_T$ for every $\zbf\in Z$ and every $\tilde t\in\{1,\ldots, T\}$ is bounded by
\begin{align}\label{nonconine}
\E[|\loss(\wsbf_T;\zbf)-\loss(\wsbf'_T;\zbf)|]\leq \frac{\tilde t M}{n}+L\E[\delta_T|\delta_{\tilde t}=0].
\end{align} This follows from the argument for a similar claim in \citep{Hardt} and applying it to our expression of~\ref{earlyupdate} parameter update.  

Now, let us define $\Delta_{t,\tilde t}\defeq \E[\delta_t|\delta_{\tilde t}=0]$. Our goal is to find an upper bound on $\Delta_{T,\tilde t}$ and then minimize it over $\tilde t$. %

At step $t$, with probability $1-1/n$, the example is the same in both $\Sc$ and $\Sc'$. Hence, we have
\begin{align}\label{ncstineq1}
\begin{split}
\delta_{t+1}&=\|(1+\mu_t)(\wsbf_t-\wsbf'_t)-{\mu_{t}}(\wsbf_{t-1}-\wsbf'_{t-1})-\alpha_t\phi_1\|\\
&\leq (1+\mu_t)\|\wsbf_t-\wsbf'_t\|+\mu_{t}\|\wsbf_{t-1}-\wsbf'_{t-1}\|+\alpha_t\|\phi_1\|\\
&\leq (1+\mu_t+\alpha_t\beta)\delta_t+\mu_{t}\delta_{t-1}
\end{split}
\end{align} where $\phi_1=\nabla_{\wbf}\loss(\wsbf_t;\zbf_{\isf_t})-\nabla_{\wbf}\loss(\wsbf'_t;\zbf_{\isf_t})$. Note that the last inequality in \eqref{ncstineq1} holds due to the $\beta$-smooth property.
With probability $1/n$, the selected example is different in $\Sc$ and $\Sc'$. In this case, we have
\begin{align}\label{ncstineq2}
\begin{split}
\delta_{t+1}&=\|(1+\mu_t)(\wsbf_t-\wsbf'_t)-\mu_{t}(\wsbf_{t-1}-\wsbf'_{t-1})-\alpha_t\phi_2\|\\
&\leq (1+\mu_t)\delta_t+\mu_{t}\delta_{t-1}+\alpha_t\|\nabla_{\wbf}\loss(\wsbf_t;\zbf_{\isf_t})\|+\alpha_t\|\nabla_{\wbf}\loss(\wsbf'_t;\zbf'_{\isf_t})\|\\
&\leq (1+\mu_t)\delta_t+\mu_{t}\delta_{t-1}+2\alpha_t L
\end{split}
\end{align} where $\phi_2=\nabla_{\wbf}\loss(\wsbf_t;\zbf_{\isf_t})-\nabla_{\wbf}\loss(\wsbf'_t;\zbf'_{\isf_t})$.

After taking expectation, for every $t\geq \tilde t$, we have 
\begin{align}\nn
\Delta_{t+1,\tilde t}\leq \big(1+\mu_t+(1-1/n)\alpha_t\beta\big)\Delta_{t,\tilde t}+\mu_{t}\Delta_{t-1,\tilde t}+2\alpha_t L/n.
\end{align}

Let us consider the recursion 
\begin{align}\nn
\tilde\Delta_{t+1,\tilde t}=\big(1+\mu_t+(1-1/n)\alpha_t\beta\big)\tilde\Delta_{t,\tilde t}+\mu_{t}\Delta_{t-1,\tilde t}+2\alpha_t L/n.
\end{align}

Note that we have $\tilde\Delta_{t+1,\tilde t}\geq \tilde\Delta_{t,\tilde t}$. Then, we have the following inequality:
\begin{align}\nn
\tilde\Delta_{t+1,\tilde t}\leq\big(1+2\mu_t+(1-1/n)\alpha_t\beta\big)\tilde\Delta_{t,\tilde t}+\frac{2\alpha_t L}{n}.
\end{align}
 Noting that $\tilde\Delta_{t,\tilde t}\geq \Delta_{t,\tilde t}$ for all $t\geq \tilde t$, we have $\E[\Delta_{T,\tilde t}]\leq S_3+S_4$ where
\begin{align}\nn
S_{3}=\sum_{t=\tilde t+1}^{t_d}\prod_{p=t+1}^{T}\Big(1+2\mu_p+\big(1-\frac{1}{n}\big)\frac{\alpha_0\beta}{p}\Big)\frac{2\alpha_0 L}{nt}
\end{align} and
\begin{align}\nn
S_{4}=\sum_{t=t_d+1}^{T}\prod_{p=t+1}^{T}\Big(1+2\mu_p+\big(1-\frac{1}{n}\big)\frac{\alpha_0\beta}{p}\Big)\frac{2\alpha_0 L}{nt}.
\end{align}
Substituting $\mu_p=\mu_d$ for $p=1,\ldots,t_d$, we can find an upper bound on $S_3$ as follows:
\begin{align}\label{S3UB}\textstyle
S_{3}&=\sum_{t=\tilde t+1}^{t_d}\prod_{p=t+1}^{T}\Big(1+2\mu_p+\big(1-\frac{1}{n}\big)\frac{\alpha_0\beta}{p}\Big)\frac{2\alpha_0 L}{nt}\nn\\
&\leq \sum_{t=\tilde t+1}^{t_d}\prod_{p=t+1}^{T}\exp\Big(2\mu_p+\big(1-\frac{1}{n}\big)\frac{\alpha_0\beta}{p}\Big)\frac{2\alpha_0 L}{nt}\nn\\
&\leq \sum_{t=\tilde t+1}^{t_d}\exp\Big(2\mu_d(t_d-t)+\big(1-\frac{1}{n}\big)\alpha_0\beta\ln\big(\frac{T}{t}\big)\Big)\frac{2\alpha_0 L}{nt}\nn\\
&\leq \frac{2\alpha_0L}{n}T^{(1-\frac{1}{n})\alpha_0\beta}\exp(2\mu_dt_d)\int_{\tilde t}^{t_d}h_1(t)t^{-(1-\frac{1}{n})\alpha_0\beta}\ud t\nn\\
&\leq \frac{2\alpha_0L}{n}T^{(1-\frac{1}{n})\alpha_0\beta}\exp(2\mu_dt_d)\int_{\tilde t}^{t_d}h_1(t)\ud t\nn\\
&=\frac{2\alpha_0L}{n}T^{(1-\frac{1}{n})\alpha_0\beta}\exp(2\mu_dt_d)\big(E_1(2\mu_d\tilde t)-E_1(2\mu_d t_d)\big)\nn
\end{align} where $h_1(t)= \frac{\exp(-2\mu_d t)}{t}$ and the exponential integral function $E_1$ is defined as
\begin{align}
E_1(x)\defeq\int_{x}^{\infty}\frac{\exp(-t)}{t}\ud t.
\end{align}

Note that the following inequalities hold for the exponential integral function for $t>0$ \citep{E1}:
\begin{align}\label{E1ineq}\textstyle
\frac{1}{2}\exp(-t)\ln\left(1+\frac{2}{t}\right)<E_1(t)< \exp(-t)\ln\left(1+\frac{1}{t}\right).
\end{align}
Applying both upper bound and lower bound in \eqref{E1ineq}, we have
\begin{align}
S_{3}\leq \frac{2\alpha_0L}{n}T^{(1-\frac{1}{n})\alpha_0\beta}h(\mu_d,t_d)
\end{align} where $h(\mu_d,t_d)=\exp\big(2\mu_d(t_d-\tilde t)\big)\ln\big(1+\frac{1}{2\mu_d\tilde t}\big)-\frac{1}{2}\ln\big(1+\frac{1}{\mu_d t_d}\big)$. 

We can also find an  upper bound on $S_4$ as follows:
\begin{align}\label{S4UB}
\begin{split}
S_{4}&=\sum_{t=t_d+1}^{T}\prod_{p=t+1}^{T}\Big(1+\big(1-\frac{1}{n}\big)\frac{\alpha_0\beta}{p}\Big)\frac{2\alpha_0 L}{nt}\\
&\leq \frac{2L}{\beta(n-1)}\Big(\frac{T}{t_d}\Big)^{(1-\frac{1}{n})\alpha_0\beta}\\
&\leq \frac{2L}{\beta(n-1)}\Big(\frac{T}{\tilde t}\Big)^{(1-\frac{1}{n})\alpha_0\beta}.
\end{split}
\end{align}

Replacing $\Delta_{T,\tilde t}$ with its upper bound in \cref{nonconine}, we obtain \cref{earlymstabnonc}.
\epr 

\cref{thm:earlymnconstab} suggests that the stability bound decreases inversely with the size of the training set. It increases as the momentum parameter $\mu_d$ increases.  {By setting $t_d=\tilde t$ in~\cref{thm:earlymnconstab} and comparing with \citep[Theorem 3.12]{Hardt},  we note that we slightly improve the exponent of $T$.  We can also establish a simpler but looser bound by noting $\tilde h(\mu_d,t_d)<h(\mu_d,t_d)=\exp\big(2\mu_d(t_d-\tilde t)\big)\ln\big(1+\frac{1}{2\mu_d\tilde t}\big)-\frac{1}{2}\ln\big(1+\frac{1}{\mu_d t_d}\big)$.}

\begin{remark}
We can show that our stability bound in~\cref{thm:earlymnconstab} holds for the projected~\ref{earlyupdate} since Euclidean projection does not increase the distance between projected points.
\end{remark}

\bcr\label[corollary]{cr:earlymstab1}

{ For~\ref{earlyupdate} with the step-size $\alpha_t=\alpha_0/t$, suppose we set $t_d=\tilde t^*+K$ where $\tilde t^*=\big(\frac{2\alpha_0L^2}{M}\big)^{\frac{1}{u+1}}T^{\frac{u}{u+1}}$ for some constant $K$. Provided that $\alpha_0\beta<1$, i.e., $u<1$, the generalization error of~\ref{earlyupdate} for $T$ steps with  $\alpha_t=\alpha_0/t$ is upper bounded by $\Oc\Big(\frac{\exp(\mu_d)T^u}{n}\Big)$, and  the number of stochastic gradient steps can grow as $n^l$ for a small $l>1$ while still allowing $\epsilon_s\rightarrow 0$ as $n\rightarrow\infty$.}
\ecr
\bpr  Note that we can minimize the expression $\frac{\tilde t M}{n}+\frac{2L^2}{\beta(n-1)}\Big(\frac{T}{\tilde t}\Big)^{u}$ in~\cref{earlymstabnonc} by optimizing $\tilde t$, where the optimal $\tilde t$ is given by $\tilde t^*$ as defined in the theorem statement. After substituting the optimal $\tilde t^*$ into~\cref{earlymstabnonc} and setting $t_d=\tilde t^*+K$ for some constant $K$, we obtain
\begin{align}\label{earlymstabnonc2}
\epsilon_s\leq\frac{2\alpha_0L^2}{n}T^{u}\chi_1+\frac{1+\frac{1}{\alpha_0\beta}}{n-1}(2\alpha_0 L^2)^{\frac{1}{u+1}}(MT)^{\frac{u}{u+1}}
\end{align} where $\chi_1=\exp(2\mu_d K)\ln\big(1+\frac{1}{2\mu_d\tilde t^*}\big)-\frac{1}{2}\ln\big(1+\frac{1}{\mu_d (\tilde t^*+K)}\big)$.

{ Note that by substituting $\tilde t=T$ into~\cref{nonconine}, for any training algorithm, we have 
\begin{align}\label{Toverngen}
\begin{split}
\E[|\loss(\wsbf_T;\zbf)-\loss(\wsbf'_T;\zbf)|]&\leq \frac{T M}{n}+L\E[\delta_T|\delta_{T}=0]\\
&=\frac{T M}{n}.  
\end{split}
\end{align}

Combining the above bounds, an upper bound on the generalization error of~\ref{earlyupdate} is given by
$\Oc\Big(\min\Big\{\frac{\exp(\mu_d)T^u}{n},\frac{TM}{n}\Big\}\Big)$. Here, we consider a nontrivial case where $u$ is small enough, i.e., the generalization error is bounded by the first term.}
\epr

{We obtain $\tilde t^*$ in~\cref{cr:earlymstab1} by optimizing over the second and third terms of the upper bound in~\cref{thm:earlymnconstab}.}

{ \paragraph{High-probability bounds.} In~\cref{app:highprob}, we establish high-probability bounds for generalization error of~\ref{earlyupdate} along the lines of~\citep{feldman2018generalization}.}

To complete our generalization analysis, in the following, we further show that~\ref{earlyupdate} updates may not satisfy uniform stability depending on how $t_d$ is set.
\bcr\label[corollary]{cr:earlymstab2}
Suppose, in~\cref{thm:earlymnconstab}, we set $t_d=\rho T$ and $\tilde t=\rho T-K$ for some $0<\rho\leq 1$ and $K<\rho T$. Then~\ref{earlyupdate} updates do not satisfy uniform stability for multiple epochs $T=\kappa n$ and the asymptotic upper bound on the penalty of generalization error is given by $\rho \kappa M$, \ie
\begin{align}\nn
\lim_{n\rightarrow\infty:T=\kappa n}\epsilon_g\leq \rho \kappa M.
\end{align}

\ecr

\bpr Substituting $t_d=\rho T$ and $\tilde t=\rho T-K$ into~\cref{earlymstabnonc}, we obtain
\begin{align}\label{earlymstabnoncpenalty}
\epsilon_s\leq\frac{2\alpha_0L^2}{n}T^{u}\chi_2+\frac{(\rho T-K) M}{n}+\frac{2L^2}{\beta(n-1)}\Big(\frac{T}{\rho T-K}\Big)^{u}
\end{align} where 
\begin{align}\nn
\chi_2=\exp\big(2\mu_d K\big)\ln\big(1+\frac{1}{2\mu_d(\rho T-K)}\big)-\frac{1}{2}\ln\big(1+\frac{1}{\mu_d \rho T}\big).
\end{align}

We can derive the asymptotic penalty by substituting $T=\kappa n$ into the upper bound \eqref{earlymstabnoncpenalty}, letting $n\rightarrow\infty$, and using~\cref{genstabth}.
\epr 

\cref{cr:earlymstab2} suggests that increasing $t_d$ worsens the generalization penalty when $t_d$ is linear in $T$. Furthermore, increasing $T$ improves the convergence bound. However, the stability upper bound increases as $T$ increases, which is expected.

{ Let $q\in [\frac{1}{2},1)$. In the following, we  establish an upper bound on the generalization error of~\ref{earlyupdate} for a general step-size $\alpha_t=\alpha_0/t^q$, which includes as special cases popular choices of step-sizes whose convergence are studied in the optimization literature~\citep{bubeck2015convex}. 

\bth\label{thm:generallrstab}
Let $q\in [\frac{1}{2},1)$. Suppose that  $\loss$ satisfies~\cref{assu:LipSm} and that  the~\ref{earlyupdate} update is executed for $T$ steps with step-size $\alpha_t=\alpha_0/t^q$ and some constant {$\mu_d\in(0,1]$} in the first $t_d$ steps. Then, for any $1 \leq \tilde t \leq t_d \leq T$,~\ref{earlyupdate} satisfies $\epsilon_s$-uniform stability with
\begin{align}\label{generallrstab}
\!\!\!\!\!\epsilon_s\leq\frac{\alpha_0L^2\sqrt{\pi}}{n\sqrt{2\mu_d}(1-q)}\exp(u_qT^{1-q})\breve h(\mu_d,t_d)+\frac{\tilde t M}{n}+\frac{2L^2}{\beta(n-1)}\exp\Big(u_q\big(T^{1-q}-{\tilde t}^{1-q}\big)\Big)
\end{align}
where $\breve h(\mu_d,t_d)=\exp(2\mu_d t_d+u_q^2/(8\mu_d))\big(\Phi(\sqrt{2\mu_d}(t_d^{1-q}+\frac{u_q}{4\mu_d}))-\Phi(\sqrt{2\mu_d}({\tilde t}^{1-q}+\frac{u_q}{4\mu_d}))\big)$,  $\Phi(x)=\erf(x):=\frac{2}{\sqrt{\pi}}\int_{0}^{x}\exp(-t^2)\ud t$,  $u_q=(1-\frac{1}{n})\frac{\alpha_0\beta}{1-q}$, and $M=\sup_{\wbf,\zbf}\loss(\wbf;\zbf)$. 
\eth}

{ \bpr Similar to the proof of~\cref{thm:earlymnconstab}, we have the following inequality:
\begin{align}\nn
\tilde\Delta_{t+1,\tilde t}\leq\big(1+2\mu_t+(1-1/n)\alpha_t\beta\big)\tilde\Delta_{t,\tilde t}+\frac{2\alpha_t L}{n}.
\end{align}
 Noting that $\tilde\Delta_{t,\tilde t}\geq \Delta_{t,\tilde t}$ for all $t\geq \tilde t$, we have $\E[\Delta_{T,\tilde t}]\leq S_3+S_4$ where
\begin{align}\nn
S_{3}=\sum_{t=\tilde t+1}^{t_d}\prod_{p=t+1}^{T}\Big(1+2\mu_p+\big(1-\frac{1}{n}\big)\frac{\alpha_0\beta}{p^q}\Big)\frac{2\alpha_0 L}{n t^q}
\end{align} and
\begin{align}\nn
S_{4}=\sum_{t=t_d+1}^{T}\prod_{p=t+1}^{T}\Big(1+2\mu_p+\big(1-\frac{1}{n}\big)\frac{\alpha_0\beta}{p^q}\Big)\frac{2\alpha_0 L}{n t^q}.
\end{align}
Substituting $\mu_p=\mu_d$ for $p=1,\ldots,t_d$, we can find an upper bound on $S_3$ as follows:
\begin{align}%
S_{3}&=\sum_{t=\tilde t+1}^{t_d}\prod_{p=t+1}^{T}\Big(1+2\mu_p+\big(1-\frac{1}{n}\big)\frac{\alpha_0\beta}{p^q}\Big)\frac{2\alpha_0 L}{n t^q}\nn\\
&\leq \sum_{t=\tilde t+1}^{t_d}\prod_{p=t+1}^{T}\exp\Big(2\mu_p+\big(1-\frac{1}{n}\big)\frac{\alpha_0\beta}{p^q}\Big)\frac{2\alpha_0 L}{n t^q}\nn\\
&\leq \sum_{t=\tilde t+1}^{t_d}\exp\Big(2\mu_d(t_d-t)+u_q\big(T^{1-q}-t^{1-q}\big)\Big)\frac{2\alpha_0 L}{n t^q}\nn\\
&\leq \frac{2\alpha_0L}{n}\exp(u_q T^{1-q}+2\mu_dt_d)\int_{\tilde t}^{t_d}\frac{\exp(-2\mu_d t-u_q t^{1-q})}{t^q}\ud t\nn\\
&\leq \frac{2\alpha_0L}{n}\exp(u_q T^{1-q}+2\mu_dt_d)\int_{\tilde t}^{t_d}\frac{\exp(-2\mu_d t^{2(1-q)}-u_q t^{1-q})}{t^q}\ud t\nn\\
&\leq \frac{2\alpha_0L}{n}\exp(u_qT^{1-q}+2\mu_dt_d+u_q^2/(8\mu_d))\int_{\tilde t}^{t_d}\frac{\exp\big(-2\mu_d(t^{1-q}+u_q/(4\mu_d))^2\big)}{t^q}\ud t\nn\\
&=\frac{\alpha_0L\sqrt{\pi}}{n\sqrt{2\mu_d}(1-q)}\exp(u_q T^{1-q}+2\mu_dt_d+u_q^2/(8\mu_d))\cdot\nn\\
&\quad\cdot\big(\Phi(\sqrt{2\mu_d}(t_d^{1-q}+\frac{u_q}{4\mu_d}))-\Phi(\sqrt{2\mu_d}(\tilde t^{1-q}+\frac{u_q}{4\mu_d}))\big)\nn
\end{align} where the fifth step holds since $2(1-q)<1$ and the last inequality follows~\citep[Eq. 3.321]{IntTable}.

We can also find an  upper bound on $S_4$ as follows:
\begin{align}\label{S4UBgenerallr}
\begin{split}
S_{4}&=\sum_{t=t_d+1}^{T}\prod_{p=t+1}^{T}\Big(1+\big(1-\frac{1}{n}\big)\frac{\alpha_0\beta}{p^q}\Big)\frac{2\alpha_0 L}{n t^q}\\
&\leq\frac{2\alpha_0 L}{n}\sum_{t=t_d+1}^{T}\exp\Big(u_q\big(T^{1-q}-t^{1-q}\big)\Big)t^{-q}\\
&\leq \frac{2L}{\beta(n-1)}\exp\Big(u_q\big(T^{1-q}-t_d^{1-q}\big)\Big)\\
&\leq \frac{2L}{\beta(n-1)}\exp\Big(u_q\big(T^{1-q}-{\tilde t}^{1-q}\big)\Big).
\end{split}
\end{align}

Replacing $\Delta_{T,\tilde t}$ with its upper bound in~\cref{nonconine}, we obtain~\cref{generallrstab}.

By its definition, we have $\Phi(x)\leq 1$. We also note that $1-\exp(-x^2)\leq \Phi(x)$ for $x>0$ following the upper bound developed for $1-\erf$ in~\citep{chiani2003new}. Applying both lower bound and upper bound on $\Phi$ in~\cref{generallrstab} and after rearranging the terms, we have
\begin{align}\label{generallrstabsim}
\epsilon_s\leq\Big(\frac{\alpha_0L^2\sqrt{\pi}}{n\sqrt{2\mu_d}(1-q)}\exp\big(2\mu_d(t_d-{\tilde t}^{2(1-q)})\big)
+\frac{2L^2}{\beta(n-1)}
\Big)\exp\Big(u_q\big(T^{1-q}-{\tilde t}^{1-q}\big)\Big)+\frac{\tilde t M}{n}.
\end{align}
\epr}

We now find a simpler expression for the generalization bound in~\cref{thm:generallrstab} by substituting $t_d$ and optimizing over $\tilde t$.  

\bcr\label[corollary]{cr:generallr}  Let $q\in [\frac{1}{2},1)$. For~\ref{earlyupdate} with a general step-size $\alpha_t=\alpha_0/t^q$, suppose we set $t_d={\tilde t^*}^{2(1-q)}+K$ for some constant $K$ where $\tilde t^*$ satisfies~\cref{opt_tilde_generallr}.
 Then the generalization error of~\ref{earlyupdate} for $T$ steps with  $\alpha_t=\alpha_0/t^q$ is upper bounded by $\Oc\Big(\min\Big\{\frac{\exp(uT^{1-q}/(u+1)+\mu_d)}{n},\frac{TM}{n}\Big\}\Big)$. 
\ecr
\bpr  Note that we can minimize: 
\begin{align}\nn
\min_{1\leq \tilde t\leq t_d}\frac{\tilde t M}{n}+\Big(\frac{\alpha_0L^2\sqrt{\pi}}{n\sqrt{2\mu_d}(1-q)}\exp\big(2\mu_dK\big)
+\frac{2L^2}{\beta(n-1)}
\Big)\exp\Big(u_q\big(T^{1-q}-{\tilde t}^{1-q}\big)\Big)    
\end{align}  
by optimizing $\tilde t$ after setting $t_d=\tilde t+K$ where the objective is the upper bound in~\cref{generallrstab}. We note that an optimal $\tilde t^*$ satisfies
\begin{align}\label{opt_tilde_generallr}
M\exp(u_q{\tilde t^*}^{1-q}){\tilde t^*}^q=\Big(\frac{u_q\alpha_0L^2\sqrt{\pi}}{\sqrt{2\mu_d}}+2L^2\alpha_0\Big)\exp(u_qT^{1-q}).\end{align}

Note \cref{opt_tilde_generallr}  does not have an analytic solution but can be solve numerically. Instead, we consider a suboptimal solution by taking $\ln$ on both sides of~\cref{opt_tilde_generallr} and applying the well-known inequality $\ln(x+1)\leq x$, $\forall x\geq -1$, which leads to:

\begin{align}\label{subopt_tilde_generallr}
{\tilde t}^{1-q} = \frac{\ln\Big(\big(\frac{u\alpha_0L^2\sqrt{\pi}}{\sqrt{2\mu_d}}+L^2\alpha_0\big)/M\Big)}{u+1}+\frac{u}{u+1}T^{1-q}.
\end{align}
 
Substituting~\cref{subopt_tilde_generallr} into~\cref{generallrstab} and combining with the upper bound in~\cref{Toverngen} complete the proof.
\epr 

As an important special case the problem considered in~\cref{thm:generallrstab}, we provide an uppe-bound on the generalization error of~\ref{earlyupdate} with the larger step size $\alpha_t=\alpha_0/\sqrt{t}$, which is a common choice in the optimization literature~\citep{bubeck2015convex}. See~\cref{app:sqrt} for the exact expression of $\tilde t$. 

\bcr\label[corollary]{cr:sqrtlr} For~\ref{earlyupdate} with the step-size $\alpha_t=\alpha_0/\sqrt{t}$, suppose we set $t_d=\tilde t^*+K$ for some constant $K$ under an optimized $\tilde t^*$, which satisfies~\cref{opt_tilde_generallr} with $q=\frac{1}{2}$. Then the generalization error of~\ref{earlyupdate} for $T$ steps with  $\alpha_t=\alpha_0/\sqrt{t}$ is upper bounded by $\Oc\Big(\min\Big\{\frac{\exp(u\sqrt{T}/(u+1)+\mu_d)}{n},\frac{TM}{n}\Big\}\Big)$. 
\ecr 

The bounds in~\cref{cr:generallr,cr:sqrtlr} complement the  results of~\citet{attia2021instability} by showing exponential growth in uniform stability bounds for stochastic gradient descent with heavy-ball momentum for possibly nonconvex problems under nonconstant step-sizes. The bound in~\cref{cr:sqrtlr} shows that as long as $T = o(\log(n)^{1/(1-q)})$,~\ref{earlyupdate} is guaranteed to generalize.

\begin{remark}[Stability of~\ref{earlyupdate} does not follow that of SGD]
As shown in~\cref{thm:exampleSGDEM} and~\cref{cr:earlymstab1}, $t_d$ can grow with $T$, \ie two iterative sequences of rules with the same starting points on two neighboring datasets can be possibly arbitrarily far after applying momentum for $t_d$ iterations. Then even a contraction map does not make the algorithm stable.  In other words, the stability of~\ref{earlyupdate} does not directly follow the stability of SGD.
\end{remark}

\section{Strongly Convex Loss}\label{sec:sconvex}
While we have discussed in the previous section the generalization of~\ref{earlyupdate} for {smooth Lipschitz loss} functions, in this section, we focus on the important class of strongly convex loss functions. We show that it suffices to consider the case $t_d=T$, \ie where~\ref{earlyupdate} becomes~\ref{update}, to achieve generalization.

\begin{assumption}[Strong convexity]\label{assu:SC}
Let $\zbf\in\Zc$ and $\ubf,\vbf\in\Omega$. The  loss function $\loss(\cdot;\zbf)$ is $\gamma$-strongly convex: there exists $\gamma>0$ such that 
\begin{align}\nn
\loss(\ubf;\zbf)\geq \loss(\vbf;\zbf)+\nabla_{\wbf} \loss(\vbf;\zbf)^\top(\ubf-\vbf)+\frac{\gamma}{2}\|\ubf-\vbf\|^2.
\end{align}
\end{assumption}

An example for $\gamma$-strongly convex loss function is Tikhonov regularization, where the empirical risk is given by  $R_\Sc(\wbf)=\sum_{i=1}^n\loss(\wbf;\zbf_i)+\frac{\gamma}{2}\|\wbf\|^2$ with a convex $\loss(\cdot;\zbf)$ for all $\zbf$. In the following, we assume that $\loss(\wbf;\zbf)$ is a $\gamma$-strongly convex function of $\wbf$ for all $\zbf\in \Zc$.

To satisfy the $L$-Lipschitz property of the loss function, we further assume that the parameter space $\Omega$ is a compact and convex set.  Since $\Omega$ is compact, the~\ref{update} update has to involve projection. %

We present a bound on the generalization of~\ref{projupdate} for $\gamma$-strongly convex loss.
\bth\label{thm:sconstab}
Suppose that  $\loss$ satisfies~\cref{assu:LipSm,assu:SC} and that~\ref{projupdate} is executed for $T$ steps with constant step-size $\alpha$ and momentum $\mu$. Provided that $\frac{\alpha\beta\gamma}{\beta+\gamma}-\frac{1}{2}\leq\mu< \frac{\alpha\beta\gamma}{3(\beta+\gamma)}$ and $\alpha\leq\frac{2}{\beta+\gamma}$,~\ref{projupdate} satisfies $\epsilon_s$-uniform stability where
\begin{align}\label{epss}
\epsilon_s\leq \frac{2\alpha L^2(\beta+\gamma)}{n\big(\alpha\beta\gamma-3\mu(\beta+\gamma)\big)}.
\end{align}
\eth

Proof sketch: the update rule in the strongly convex case is a contraction, which is not the case in the convex case. In particular, the contraction term due to  $\gamma$-strong convexity can be leveraged to control the additional expansion term due to momentum. The overall update remains a contraction assuming the momentum is not too large.  See~\cref{app:thm:sconstab} for the complete proof.

\cref{thm:sconstab} implies that the stability bound decreases inversely with the size of the training set. It increases as the momentum parameter $\mu$ increases. These properties are also verified in our experimental evaluation.\footnote{Our purpose in this work is \textit{not} to show the superiority of~\ref{update} or~\ref{earlyupdate}, in terms of the \textit{stability bound}, over SGD. Given the known advantage of~\ref{update} in terms of speeding up training, our purpose is to further analyze the stability/generalization properties of~\ref{update} and~\ref{earlyupdate}.}

The theoretically advocated momentum parameters in \citep{Polyak,Nesterov} are based on \textit{convergence} analysis of gradient descent with momentum, and do not account for \textit{generalization}. Depending on the condition number of the problem, these values may not satisfy the range of momentum in~\cref{thm:sconstab}. These values are not necessarily optimal for~\ref{projupdate}, in terms of our objective of true risk. Our goal in~\cref{thm:sconstab} is to show nontrivial cases that~\ref{projupdate} satisfies uniform stability. %

\begin{remark} Compared with the stability bound in \citep{Hardt} for SGD, both bounds are in $O(1/n)$. Our bound in~\cref{thm:sconstab} holds for $\alpha\leq \frac{2}{\beta+\gamma}$, which is slightly less restrictive than the range of step-size in \citep[Theorem 3.9]{Hardt}. By substituting $\mu=0$ in~\cref{epss}, we note that the constant term of our bound, $\frac{\beta+\gamma}{\beta\gamma}$, is slightly larger than that of \citep[Theorem 3.9]{Hardt}, which is ${1}/{\gamma}$. Compared with \citep{Chen},
our bound is independent of $T$ and our work analyzes the case of strongly convex loss.
\end{remark}

Classical generalization bounds using Rademacher complexity, which measures the rate of uniform convergence, are obtained for linear predictors with various norm constraints \citep{Shai2014}. Those classical generalization bounds are typically $\Oc(1/\sqrt{n})$. For linear predictors with a Lipschitz loss and a strongly convex regularizer, by bounding Rademacher complexity, it has been shown that with high probability, the generalization error is bounded by $\Oc(1/\sqrt{n})$ for all parameters in a certain bounded set \citep{Kakade2008}. { The fast rates by~\citet{sridharan2008fast} for regularized linear prediction are built based on the notion of localized Rademacher complexity~\citep{bartlett2002localized}, which requires an additional boundedness on the dual norm of data mapping.  Our high-probability generalization bound for general  smooth Lipschitz loss functions in~\cref{app:highprob} is $\Oc(1/n)$.} 

Our stability analysis captures how the learning algorithm explores the hypothesis class, in particular, how the generalization gap depends on the momentum. More broadly, unlike stability, uniform convergence is not necessary for learning (based on the learnability definition in \citep{SSSS}) in the general learning setting \citep{SSSS}.

Finally, in the case of strongly convex loss, we can further consider the minimization of the true risk as defined in~\cref{Truerisk}, since we are able to derive an upper bound on the optimization error (shown in~\cref{app:thm:sconopt}). In~\cref{app:prop:optlearning},  we study how the uniform stability results in an upper bound on the true risk of~\ref{projupdate}.  We first show that stability results similar to~\cref{thm:sconstab} hold even if the average parameter $\hat\wbf_T$ is considered as the output of algorithm $A$.  We then decompose the expected true risk into a stability error term and an optimization one. 
We also compare the final results with SGD with no momentum, and we show that one can achieve tighter bounds by using~\ref{projupdate} than vanilla SGD.

\section{Experimental Evaluation}\label{sec:numerical}
\subsection{Nonconvex Loss}
\begin{figure*}[t]
    \includegraphics[width=.32\textwidth]{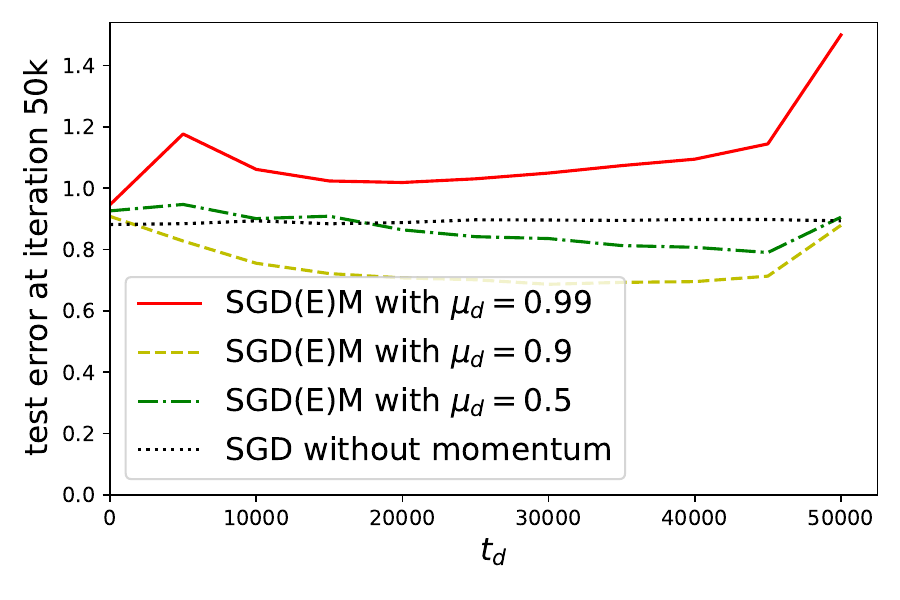}
    \hfill
    \includegraphics[width=.32\textwidth]{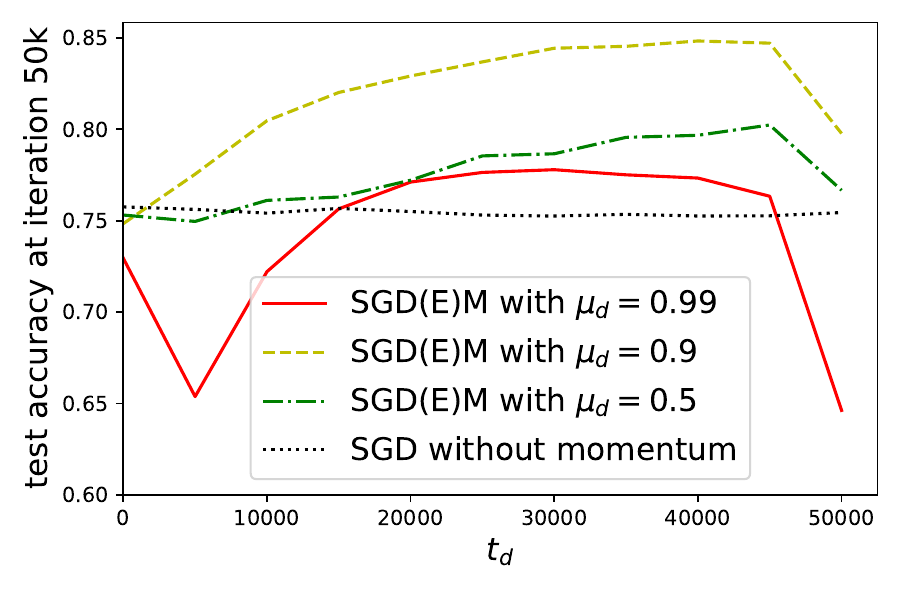}
    \hfill
    \includegraphics[width=.32\textwidth]{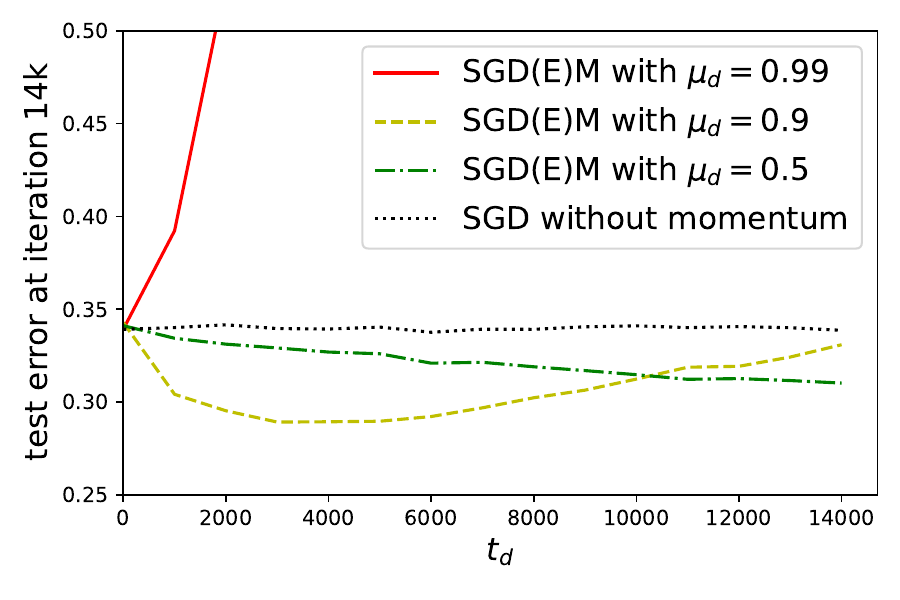}
    \caption{Test error (left) and test accuracy (middle) of ResNet-20 on CIFAR10. Test error of a feedforward fully connected neural network for notMNIST dataset (right).  }
    \label{fig:nonconv}
\end{figure*}

In this section, we validate the insights obtained in our theoretical results using experimental evaluation. Our main goal is to study how adding momentum affects the generalization and convergence of SGD.

We first investigate the performance of~\ref{earlyupdate} when applied to both CIFAR10 \citep{CIFAR10} and notMINIST datasets for nonconvex loss functions. We set $T$ to 50000 and 14000 for CIFAR10 and notMNIST experiments, respectively. For each value of $\mu_d$, we add momentum for 0-10 epochs. For each pair of $(\mu_d,t_d)$, we repeat the experiments 10 times with random initializations. \ref{update} can be viewed as a special form of~\ref{earlyupdate} when the momentum is added for the entire training (\ie $t_d=T$). For 10 epochs and without data augmentation, we train ResNet-20 on CIFAR10  and a feedforward fully connected neural network with 1000 hidden nodes on notMINIST. For the feedforward fully connected neural network, we use ReLU activation functions, a cross-entropy loss function, and a softmax output layer with Xavier initialization to initialize the weights \citep{Xavier}.\footnote{We observe similar results for smooth activation functions.} We set the step-size $\alpha=0.01$. The minibatch size is set to 10. We use 10~\eqref{earlyupdate} realizations to evaluate the average performance. We compare the test performance of SGD without momentum with that of~\ref{earlyupdate} under $\mu_d=0.5$, $\mu_d=0.9$, and $\mu_d=0.99$.

\textbf{Outperforming both SGD and~\ref{update}.} We show the test error and test accuracy versus $t_d$ under~\ref{earlyupdate} for CIFAR10 dataset in~\cref{fig:nonconv} (left and middle).
We observe that adding momentum for the entire training (\ie $t_d=T$ or~\ref{update}) is useful when the momentum parameter is small. For different $\mu_d$ values, we notice there exists an optimal $t_d$ in~\cref{fig:nonconv} (left) when test error is minimized. We plot the test error versus $t_d$ for notMNIST dataset in~\cref{fig:nonconv} (right). The test accuracy is shown in~\cref{app:exp}. We observe an overshooting phenomenon for $\mu_d=0.99$, which is consistent with our convergence analysis in~\cref{thm:earlymnconconv}. {We observe similar phenomenon when we train a feedforward fully connected neural network with 1000 hidden nodes on MINIST dataset.} In terms of test accuracy, we observe that it is not helpful to use a momentum parameter, $\mu_d\approx 1$, for the entire training. In an online framework with high dimensional parameters, early momentum is particularly useful since we can minimize memory utilization as~\ref{earlyupdate} does not require $\wbf_{t-1}$ for the entire iterative updates.%

\begin{table*}[!tb]
\footnotesize
\setlength{\tabcolsep}{10pt}
\renewcommand{\arraystretch}{0.5}
\centering
\caption{ Ablation studies where we optimize performance of SGD and~\ref{update}  by obtaining the minimum test error over step-sizes $\alpha\in\{0.1, 0.01, 0.001, 0.0001\}$ and momentum parameters $\mu\in\{0,0.5,0.9,0.99\}$. We do not tune $t_d$ and used the fixed $t_d=3000$. The rest of the setup is similar to~\cref{fig:nonconv} (right).}
\label{tab:ablation}
\begin{tabular}{llll}
\toprule
{}  &   \textbf{SGD} &              \textbf{\ref{earlyupdate}} &              \textbf{\ref{update}} \\
\midrule
Test error  &  0.3596 $\pm$ 0.0142 &              {\bfseries  0.2892} $\pm$ 0.0042 &             0.3194 $\pm$ 0.0030 \\
\bottomrule
\end{tabular}
\end{table*}

{ To see whether~\ref{update} is able to match performance of~\ref{earlyupdate} by further tuning hyperparameters, in~\cref{tab:ablation}, we show the results of an ablation study where we minimize the test error of SGD and~\ref{update}  by optimizing over step-sizes $\alpha\in\{0.1, 0.01, 0.001, 0.0001\}$ and momentum parameters $\mu\in\{0,0.5,0.9,0.99\}$. For~\ref{earlyupdate}, we do~{\it not} tune $t_d$ and used the fixed $t_d=3000$. The rest of the setup is similar to~\cref{fig:nonconv} (right). We repeat the experiments for 5 times to report confidence intervals. These results show that even under tuned hyperparameters,~\ref{earlyupdate}  outperforms both~\ref{update} and SGD in terms of test error.}     

\begin{wrapfigure}{R}{0.4\textwidth}
        \vspace*{-0.7cm}
            \includegraphics[width=0.39\textwidth]
            {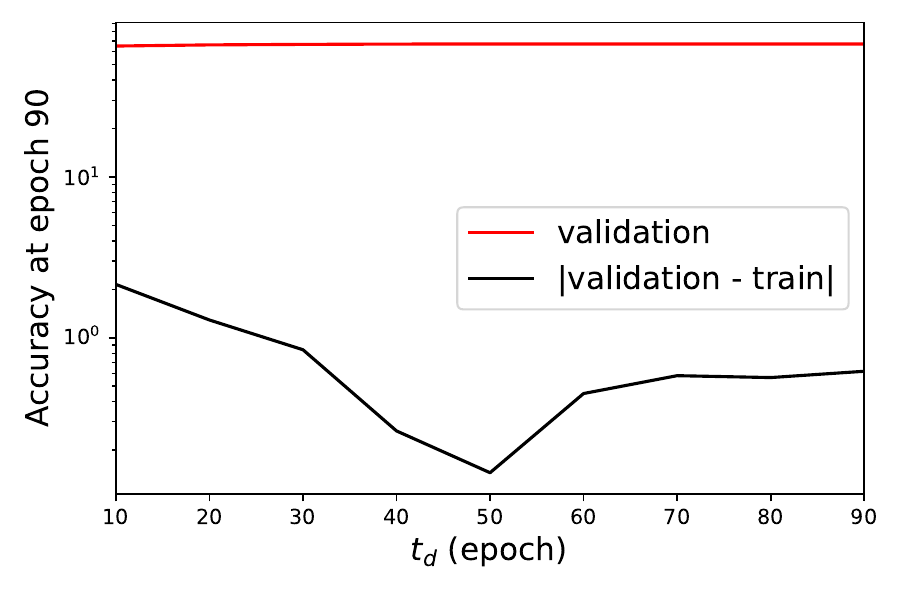}
\caption{Validation accuracy and generalization gap of~\ref{earlyupdate} when training ResNet-18 on ImageNet in a distributed setting with 4 GPUs under tuned step-size and global minibatch size of 128. For each $t_d$, the momentum is set to  $\mu_d=0.9$ in the first $t_d$ epochs and then zero for the next $90-t_d$ epochs. \ref{update} is a special form of~\ref{earlyupdate} with $t_d=90$.  }
\label{fig:ImageNet_testgenaccvst}
        \vspace*{-0.4cm}
   \end{wrapfigure}

\paragraph{Distributed training on ImageNet.} \cref{fig:ImageNet_testgenerrvst} shows validation loss and generalization error of~\ref{earlyupdate} at epoch 90 when training ResNet-18 on ImageNet in a practical data-parallel setting with 4 GPUs under \textit{tuned step-sizes} for SGD and~\ref{update}. We observe that the minimum generalization error happens if the momentum is applied for 50 epochs. In~\cref{fig:ImageNet_testgenaccvst},  we plot validation accuracy and generalization gap of~\ref{earlyupdate} at epoch 90.  Similar to the loss results, we observe that the minimum generalization error happens if the momentum is applied for 50 epochs.  Our accuracy results are on par with existing results~\citep{Resnet}.

\paragraph{Details of ImageNet experiments.} The  global minibatch size and weight decay are set to 128 and $5\times10^{-5}$, respectively. For each $t_d$, the momentum is set to  $\mu_d=0.9$ in the first $t_d$ epochs and then zero for the next $90-t_d$ epochs. We use a cluster with 4 NVIDIA 2080 Ti GPUs with the following  CPU details: Intel(R) Xeon(R) CPU E5-2650 v4 @ 2.20GHz; 48 cores; GPU2GPU bandwidth: unidirectional 10GB/s and bidirectional 15GB/s.

\subsection{Strongly Convex Loss} 
\begin{figure*}[t]
    \includegraphics[width=.32\textwidth]{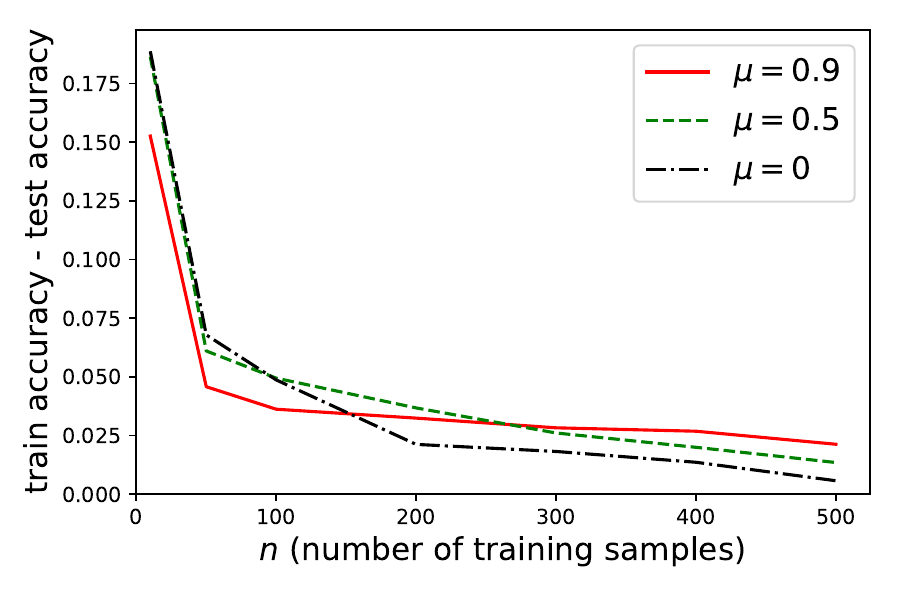}
    \hfill
    \includegraphics[width=.32\textwidth]{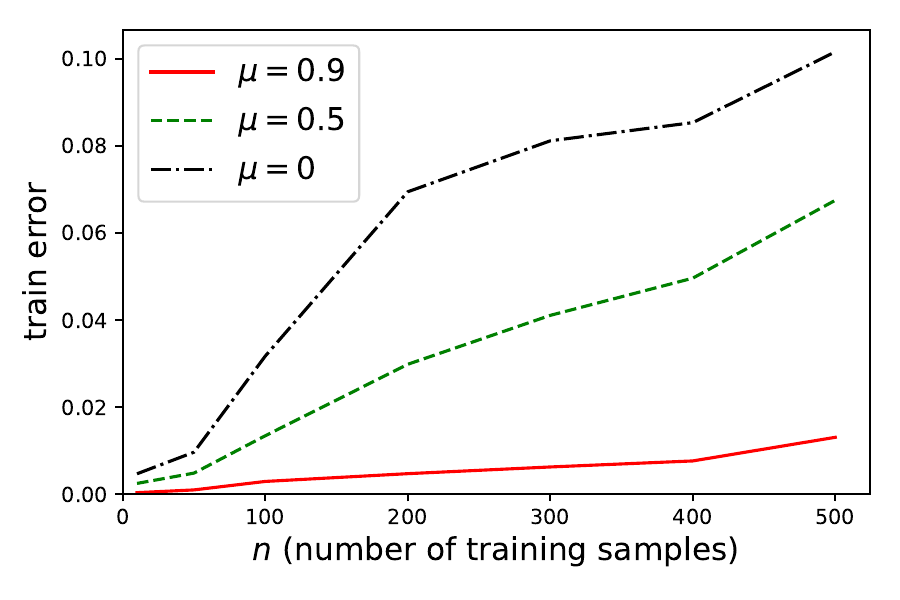}
    \hfill
    \includegraphics[width=.32\textwidth]{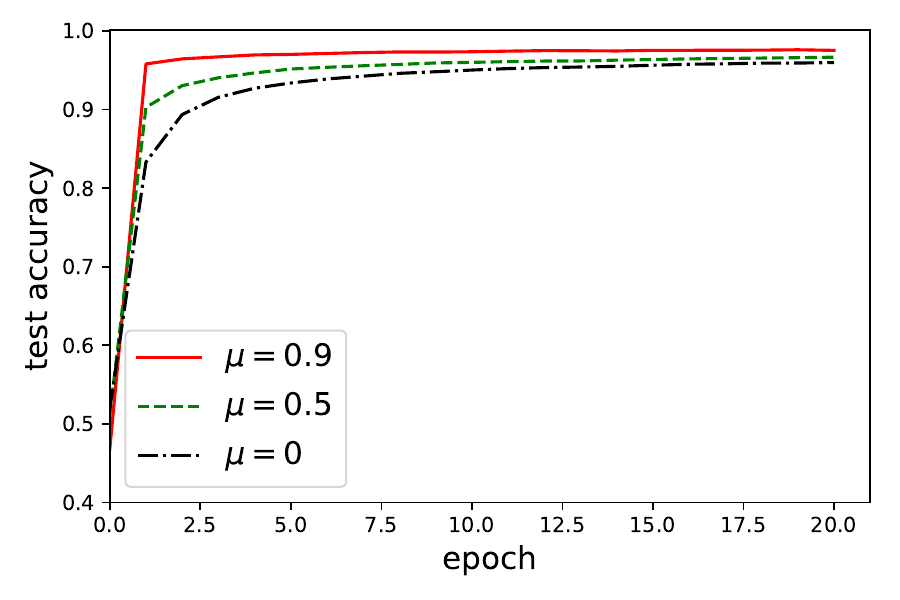}
    \caption{Generalization error (left) and training error (middle) of logistic regression (cross entropy loss) for notMNIST dataset with $T=1000$ iterations . Test accuracy of logistic regression for notMNIST dataset with $n=500$ (right).}
    \label{fig:sconv}
\end{figure*}

We now study the performance of~\ref{earlyupdate} for a smooth and strongly convex loss function. We train a logistic regression model with the weight decay regularization on notMNIST and MNIST. The setup's details are provided in~\cref{app:exp}.  We plot the test error and test accuracy versus $t_d$ under~\ref{earlyupdate} for notMNIST and MNIST in~\cref{app:exp} and observe that, unlike the case of nonconvex loss functions, it does not hurt to add momentum for the entire training. We then focus on~\ref{update} and compare the optimization and generalization performance of vanilla SGD with that of~\ref{update} under $\mu=0.5$ and $\mu=0.9$, which are common momentum values used in practice~\citep[Section 8.3.2]{DLbook}.

\paragraph{Hurting generalization error and improving training error.} In~\cref{fig:sconv} (left) and (middle), we plot generalization and training error versus $n$ with fixed $T$ and observe that generalization error decreases as $n$ increases  for all values of $\mu$, which is suggested by our stability upper bound in~\cref{thm:sconstab}.  In addition, for sufficiently large $n$, we observe that the generalization error  increases with $\mu$, consistent with~\cref{thm:sconstab}. On the other hand, training error increases as $n$ increases with fixed $T$, which is expected. We can observe that adding momentum reduces training error as it improves the convergence rate.

\paragraph{Negligible improvement of test accuracy.} In~\cref{fig:sconv} (right), we plot test accuracy versus $T$  with fixed $n$ (See~\cref{app:exp} for  training error, training accuracy, and test error). As the number of epochs increases, we note that  the benefit of  momentum on the test accuracy becomes negligible. This happens because  adding momentum results in a  higher generalization error thus penalizing the gain in training error.

\section{Conclusions and Future Work}\label{Sec:conc}
We study the generalization error  of~\ref{earlyupdate} under mild technical conditions. We show that there exists a convex loss function for which the stability gap for multiple epochs of~\ref{update} becomes unbounded and investigate a modified momentum-based update rule, \ie \ref{earlyupdate}. We establish a bound on the generalization error of~\ref{earlyupdate} for the class of {smooth Lipschitz loss} functions. Our results confirm that deep neural networks can be trained for multiple epochs of~\ref{earlyupdate} while their generalization errors are bounded. We also study the convergence of~\ref{earlyupdate} in terms of a bound on the expected norm of the gradient. Then, for the case of strongly convex loss functions, we establish an upper bound on the generalization error, which decreases with the size of the training set, and increases as the momentum parameter is increased. We establish an upper bound on the expected difference between the true risk of~\ref{projupdate} and the global minimum of the empirical risk. Finally, we present experimental evaluation and show that the numerical results are consistent with our theoretical bounds and~\ref{earlyupdate} is an effective algorithm for nonconvex problems.

Beyond uniform stability analysis, which is sufficient for generalization, developing necessary conditions for generalization of various learning algorithms remains an open problem. In particular, ``on-average'' stability is a more relaxed notion and depends on the data-generating distribution~\citep{SSSS}.  %
Finally, practical methods for adaptive training, such as Adam, use a variation of the heavy-ball momentum \citep{adam}. Adapting our analysis for such extensions is also an interesting area of future work. %

\acks{This work was supported by 1) the Research Council of Norway through its Centres of Excellence scheme, Integreat - Norwegian Centre for knowledge-driven machine learning, project number 332645; 2) the Research Council of Norway through its Centre for Research-based Innovation funding scheme (Visual Intelligence under grant no. 309439), and Consortium Partners; 3) Hasler Foundation Program: Hasler Responsible AI (project number 21043); 4) the Swiss National Science Foundation (SNSF) under grant number 200021\_205011; 5) the Natural Sciences and Engineering Research Council of Canada under a Discovery Grant.}

\newpage

\appendix

\paragraph{Content of the appendix.} The appendix is organized as follows: 
\begin{itemize}
    \item In~\cref{app:thm:exampleSGDM_lr}, we show that the stability gap for multiple epochs of~\ref{update} may become unbounded for any step-size schedule.
    \item \cref{thm:generallrconv} (convergence of~\ref{earlyupdate}) is proved in~\cref{app:thm:generallrconv}. 
    \item Convergence of~\ref{earlyupdate} with constant step-size is provided in~\cref{app:thm:earlymnconconv}.
    \item Convergence guarantees for~\ref{earlyupdate} with other time-dependent step-sizes are provided in ~\cref{app:timelr_decay}.
    \item A sufficient condition for the upper bound in~\cref{thm:earlymnconconv} to become monotonically decreasing is provided in~\cref{app:cor:SGDEMvstd}.
    \item Benefit of using momentum in terms of convergence is elaborated in ~\cref{app:momentumrole}.
    \item High-probability generalization bounds for~\ref{earlyupdate} are established in~\cref{app:highprob}. 
    \item \cref{thm:sconstab} (stability of strongly convex problems) is proved in ~\cref{app:thm:sconstab}.
    \item Convergence bound for strongly convex loss is provided in~\cref{app:thm:sconopt}.
    \item An upper bound on true risk of strongly convex loss is established in~\cref{app:prop:optlearning}.
    \item Generalization error of~\ref{earlyupdate} with  $\alpha_t=\alpha_0/\sqrt{t}$ is discussed in~\cref{app:sqrt}.
    \item Additional experimental details are included in~\cref{app:exp}.  
\end{itemize}

In our analysis of the stability of~\ref{update}, we will consider the following two properties of the growth of the update rule. Let $\Omega$ denote the model parameter space. Consider a general update rule $G$ which maps $\wbf\in\Omega$ to another point $G(\wbf)\in\Omega$. Our goal is to track the divergence of two different iterative sequences of update rules with the same starting point.
\begin{definition} An update rule $G$ is $\eta$-expansive if %
\begin{align}\nn
\sup_{\vbf,\wbf\in\Omega}\frac{\|G(\vbf)-G(\wbf)\|}{\|\vbf-\wbf\|}\leq \eta.
\end{align}
\end{definition}

\begin{definition} An update rule $G$ is $\sigma$-bounded if %
\begin{align}\nn
\sup_{\wbf\in\Omega}\|\wbf-G(\wbf)\|\leq \sigma.
\end{align}
\end{definition}

\section{Importance of early momentum on bounding the stability gap}\label{app:thm:exampleSGDM_lr}{%
To highlight the importance of early momentum on bounding the stability gap, in this section, we show that the stability gap for multiple epochs of~\ref{update} may become unbounded for any step-size schedule. This includes $\alpha_1=1$ and  $\alpha_j=0$ for $j>1$, \ie  the gradient term is added only in the first iteration. We also establish a $\Omega(\frac{T}{n})$ lower bound for~\ref{update} on~\cref{ex1} even with a \textit{time-decaying} step-size, which shows that it is important to control both step-size and momentum to establish uniform stability.}
\bth\label{thm:exampleSGDM_lr} For~\cref{ex1} with datasets described in~\cref{thm:lowerbound} and for \textit{any step-size schedule}, there exists a momentum such that the stability gap for~\ref{update} is lower bounded by $\Omega(\frac{T}{n})$. 

In addition, if $\alpha_j \geq\alpha_{\min}$ for $j = 1,2,\ldots, \eta T$ where $\alpha_{\min}$ and $\eta<1$ are some constants that do not depend on $T$, then the stability gap for~\ref{update} is lower bounded by $\Omega(\frac{T}{n})$ for \textit{any momentum} $\mu>0$. 
\eth
\bpr
Following the proofs of~\cref{thm:lowerbound} for~\ref{update} with $\alpha_1>0$, we have 
\begin{align}\nn
\E_A[|\loss(\wsf_T;\zbf)-\loss(\wsf'_T;\zbf)|]\geq\frac{2\alpha_1 L^2}{n}\sum_{j=0}^{T-1}\mu^j.
\end{align}
 Substituting $\mu= 1$, $\E_A[|\loss(\wsf_T;\zbf)-\loss(\wsf'_T;\zbf)|]$ is lower bounded by $\Omega(\frac{T}{n})$.  

For the second part of the theorem, suppose $\alpha_j \geq\alpha_{\min}$ for $j = 1,2,\ldots, \eta T$. Then we have
\begin{align}\nn
\E_A[|\loss(\wsf_T;\zbf)-\loss(\wsf'_T;\zbf)|]\geq\frac{2\alpha_{\min} L^2}{n} \sum_{j=0}^{\eta T-1}(T-j)\mu^j
\end{align}
and $\E_A[|\loss(\wsf_T;\zbf)-\loss(\wsf'_T;\zbf)|]$ is lower bounded by $\Omega(\frac{T}{n})$. 
\epr

{%
\bcr For~\cref{ex1} with datasets described in~\cref{thm:lowerbound} and for \textit{time-decaying step-size}, the stability gap for~\ref{update} is lower bounded by $\Omega(\frac{T}{n})$ for \textit{any momentum} $\mu>0$.
\ecr
\bpr It follows immediately from the proof of~\cref{thm:exampleSGDEM}. 
\epr 
}

\section{Proof of~\cref{thm:generallrconv} (convergence of~\ref{earlyupdate})}\label{app:thm:generallrconv}

 The following lemmas are useful for our proofs: 

\blm\label[lemma]{lm:dsum}
For any integers $t$, $p$, and $q$, such that $0 \leq t\leq T-1$, $p<T$, and $q<T-t$, and for any sequences 
 $a_0,a_1,\ldots$, and $b_0,b_1,\ldots$, we have 

\begin{align}\nn
    \sum_{i=p}^{T}a_i\sum_{j=q}^{i-t}b_j =  \sum_{i=q}^{T-t}b_i\sum_{j=i+t}^{T}a_j.
\end{align}
\elm
\bpr
We prove by induction. It clearly holds for $T=1$. Suppose it holds for all $k<T$. Then, we have 

\begin{align}
\begin{split}
    \sum_{i=p}^{k+1}a_i\sum_{j=q}^{i-t}b_j &= \sum_{i=p}^{k}a_i\sum_{j=q}^{i-t}b_j + a_{k+1}\sum_{j=q}^{k+1-t}b_j\\
    &= \sum_{i=q}^{k-t}b_i\sum_{j=i+t}^{k}a_j + a_{k+1}\sum_{j=q}^{k+1-t}b_j\\
    &=\sum_{i=q}^{k-t}b_i\sum_{j=i+t}^{k}a_j + a_{k+1}\sum_{j=q}^{k-t}b_j + a_{k+1}b_{k+1-t}\\
    &= \sum_{i=q}^{k-t}b_i\sum_{j=i+t}^{k+1}a_j +a_{k+1}b_{k+1-t}\\
    &=\sum_{i=q}^{k+1-t}b_i\sum_{j=i+t}^{k+1}a_j.
\end{split}
\end{align}
\epr

As two special cases of ~\cref{lm:dsum}, we obtain~\citep[Lemma 4]{li2020high} by substituting $q=1,~t=0,~p=1$ and $q=0,~t=1,~p=1$: 

\begin{align}\nn
    \sum_{i=1}^{T}a_i\sum_{j=1}^{i}b_j &=  \sum_{i=1}^{T}b_i\sum_{j=i}^{T}a_j,\\
    \sum_{i=1}^{T}a_i\sum_{j=0}^{i-1}b_j &=  \sum_{i=1}^{T-1}b_i\sum_{j=i+1}^{T}a_j.\nn
\end{align}

To facilitate the convergence analysis, we  define $\qsbf_t\defeq \wsbf_t-\wsbf_{t-1}$ with $\qsbf_0=0$ and $\qsbf_1=0$. It is not difficult to show that $\qsbf_{t+1}= \mu_d\qsbf_{t} -\alpha_t\nabla_\wbf \loss(\wsbf_t;\zbf_{\isf_t})$. Since the empirical risk $R_\Sc$ is a $\beta$-smooth function, we have
\begin{align}\label{generallrconvine1}
\begin{split}
R_\Sc(\wsbf_{t+1})&\leq R_\Sc(\wsbf_{t})+\nabla R_\Sc(\wsbf_{t})^\top\qsbf_{t+1}+\frac{\beta}{2}\|\qsbf_{t+1}\|^2.
\end{split}
\end{align}

Based on the definition of $\qsbf_t$, the inner-product term in~\cref{generallrconvine1} is bounded: 

\begin{align}\label{generallrconvinner}
\begin{split}
\nabla R_\Sc(\wsbf_{t})^\top\qsbf_{t+1} &= \mu_d\nabla R_\Sc(\wsbf_{t})^\top\qsbf_{t} -  \alpha_t\nabla R_\Sc(\wsbf_{t})^\top \nabla \loss(\wsbf_t;\zbf_{\isf_t})\\
 &= \mu_d\nabla R_\Sc(\wsbf_{t-1})^\top\qsbf_{t} 
+ \mu_d\big(\nabla R_\Sc(\wsbf_{t})-\nabla R_\Sc(\wsbf_{t-1})\big)^\top\qsbf_{t}\\
&\quad -  \alpha_t\nabla R_\Sc(\wsbf_{t})^\top \nabla \loss(\wsbf_t;\zbf_{\isf_t})\\
 &\leq \mu_d\nabla R_\Sc(\wsbf_{t-1})^\top\qsbf_{t} 
+ \mu_d\|\nabla R_\Sc(\wsbf_{t})-\nabla R_\Sc(\wsbf_{t-1})\|\|\qsbf_{t}\|\\
&\quad -  \alpha_t\nabla R_\Sc(\wsbf_{t})^\top \nabla \loss(\wsbf_t;\zbf_{\isf_t})\\
&\leq \mu_d\nabla R_\Sc(\wsbf_{t-1})^\top\qsbf_{t} 
+ \mu_d\beta\|\qsbf_{t}\|^2-  \alpha_t\nabla R_\Sc(\wsbf_{t})^\top \nabla \loss(\wsbf_t;\zbf_{\isf_t})
\end{split}
\end{align}
where the last inequality holds due to smoothness. Unraveling the recursion~\cref{generallrconvinner}, we have
\begin{align}\nn%
\nabla R_\Sc(\wsbf_{t})^\top\qsbf_{t+1}&\leq  \beta \sum_{i=0}^{t-1}\mu_d^{t-i}\|\qsbf_{i+1}\|^2-  
\sum_{i=1}^{t}\mu_d^{t-i} \alpha_i\nabla R_\Sc(\wsbf_{i})^\top \nabla \loss(\wsbf_i;\zbf_{\isf_i})
\end{align}
For simplicity of analysis, we first suppose that the momentum is applied in $T$ steps. Then we modify the bound considering it is set to  zero after $t_d$ steps. 
Substituting this bound in~\cref{generallrconvine1} and 
summing for $t=1,\ldots,T$, we have  
\begin{align}\label{generallrconvine2}
\begin{split}
R_\Sc(\wsbf_{T+1})&\leq R_\Sc(\wsbf_{0})+\beta \sum_{t=1}^T\sum_{i=1}^{t-1}\mu_d^{t-i}\|\qsbf_{i+1}\|^2+\frac{\beta}{2}\sum_{t=1}^T\|\qsbf_{t+1}\|^2\\
&\quad -  
\sum_{t=1}^T\sum_{i=1}^{t}\mu_d^{t-i} \alpha_i\nabla R_\Sc(\wsbf_{i})^\top \nabla \loss(\wsbf_i;\zbf_{\isf_i}).
\end{split}
\end{align}

Using~\cref{lm:dsum}, we expand the the second and fourth terms in the upper bound~\cref{generallrconvine2}:

\begin{align}\label{sumlm1}
\begin{split}
\beta \sum_{t=1}^T\sum_{i=1}^{t-1}\mu_d^{t-i}\|\qsbf_{i+1}\|^2
&=\beta  \sum_{t=1}^{T-1}\|\qsbf_{t+1}\|^2\mu_d^{-t} \sum_{i=t+1}^{T}\mu_d^{i}\\
&= \beta  \sum_{t=1}^{T-1}\frac{\mu_d-\mu_d^{T-t+1}}{1-\mu_d} \|\qsbf_{t+1}\|^2\\
&\leq \frac{\beta\mu_d}{1-\mu_d}  \sum_{t=1}^{T-1} \|\qsbf_{t+1}\|^2.
\end{split}
\end{align}

Furthermore, we have

\begin{align}\label{sumlm2}
\begin{split}
-  \sum_{t=1}^T\sum_{i=1}^{t}\mu_d^{t-i} \alpha_i\nabla R_\Sc(\wsbf_{i})^\top \nabla \loss(\wsbf_i;\zbf_{\isf_i})&=
- \sum_{t=1}^T\sum_{i=1}^{t}\mu_d^{t-i} \alpha_i\nabla R_\Sc(\wsbf_{i})^\top \nabla R_\Sc(\wsbf_{i})\\
&\quad + \sum_{t=1}^T\sum_{i=1}^{t}\mu_d^{t-i} \alpha_i\nabla R_\Sc(\wsbf_{i})^\top \big(\nabla R_\Sc(\wsbf_{i})-\nabla \loss(\wsbf_i;\zbf_{\isf_i})\big)\\
&=
- \sum_{t=1}^T\mu_d^{-t}\alpha_t\|\nabla R_\Sc(\wsbf_{t})\|^2\sum_{i=t}^{T}\mu_d^{i}\\
&\quad + \sum_{t=1}^T\mu_d^{-t}\alpha_t\nabla R_\Sc(\wsbf_{t})^\top \big(\nabla R_\Sc(\wsbf_{t})-\nabla \loss(\wsbf_t;\zbf_{\isf_t})\big)\sum_{i=t}^{T}\mu_d^{i}\\
&=
- \sum_{t=1}^T\frac{1-\mu_d^{T-t+1}}{1-\mu_d}\alpha_t\|\nabla R_\Sc(\wsbf_{t})\|^2\\
&\quad + \sum_{t=1}^T\frac{1-\mu_d^{T-t+1}}{1-\mu_d}\alpha_t\nabla R_\Sc(\wsbf_{t})^\top \big(\nabla R_\Sc(\wsbf_{t})-\nabla \loss(\wsbf_t;\zbf_{\isf_t})\big)
\end{split}
\end{align}

Substituting~\cref{sumlm1,sumlm2} into~\cref{generallrconvine2} and rearranging the terms, we obtain 

\begin{align}\label{generallrconvine3}
\begin{split}
\sum_{t=1}^T\frac{1-\mu_d^{T-t+1}}{1-\mu_d}\alpha_t\|\nabla R_\Sc(\wsbf_{t})\|^2&\leq R_\Sc(\wsbf_{0}) - R_\Sc(\wsbf_{T+1})+\frac{\beta}{2}\sum_{t=1}^T\|\qsbf_{t+1}\|^2\\
&\quad + \sum_{t=1}^T\frac{1-\mu_d^{T-t+1}}{1-\mu_d}\alpha_t\nabla R_\Sc(\wsbf_{t})^\top \big(\nabla R_\Sc(\wsbf_{t})-\nabla \loss(\wsbf_t;\zbf_{\isf_t})\big)\\
&\quad + \frac{\beta\mu_d}{1-\mu_d}  \sum_{t=1}^{T-1} \|\qsbf_{t+1}\|^2.
\end{split}
\end{align}
We now find an upper bound on $\sum_{t=1}^T\|\qsbf_{t+1}\|^2$: 
\begin{align}\nn
\begin{split}
\sum_{t=1}^T\|\qsbf_{t+1}\|^2&= \sum_{t=1}^T\|\mu_d\qsbf_{t} -\alpha_t\nabla_\wbf \loss(\wsbf_t;\zbf_{\isf_t})\|^2\\
&\leq \sum_{t=1}^T \mu_d \|\qsbf_{t}\|^2+
\sum_{t=1}^T \frac{1}{1-\mu_d} \|\alpha_t\nabla\loss(\wsbf_t;\zbf_{\isf_t})\|^2\\
&\leq \sum_{t=1}^{T+1} \mu_d \|\qsbf_{t}\|^2+
\sum_{t=1}^{T} \frac{1}{1-\mu_d} \|\alpha_t\nabla \loss(\wsbf_t;\zbf_{\isf_t})\|^2\\
&\leq \sum_{t=1}^{T} \mu_d \|\qsbf_{t+1}\|^2+
\sum_{t=1}^{T} \frac{1}{1-\mu_d} \|\alpha_t\nabla \loss(\wsbf_t;\zbf_{\isf_t})\|^2\\
&\leq \sum_{t=1}^{T} \frac{1}{(1-\mu_d)^2} \|\alpha_t\nabla_\wbf \loss(\wsbf_t;\zbf_{\isf_t})\|^2\\
&\leq \frac{L^2}{(1-\mu_d)^2}\sum_{t=1}^{T} \alpha_t^2.
\end{split}
\end{align}
where the second and last lines hold due to Jensen's inequality and $L$-Lipschitz property, respectively. 

Applying this upper bound in~\cref{generallrconvine3}, taking expectation over $\isf_0,\ldots,\isf_{t_d}$, we have: 
\begin{align}\label{generallrconvKey1}
\begin{split}
\E_{A}\Big[\sum_{t=1}^{t_d}\frac{1-\mu_d^{t_d-t+1}}{1-\mu_d}\alpha_t\|\nabla R_\Sc(\wsbf_{t})\|^2\Big]&\leq R_\Sc(\wsbf_{0}) - \inf_{\wbf} R_\Sc(\wbf)+\frac{\beta L^2}{2(1-\mu_d)^2}\sum_{t=1}^{t_d} \alpha_t^2\\
&\quad + \frac{\beta\mu_d L^2}{(1-\mu_d)^3}  \sum_{t=1}^{t_d-1} \alpha_t^2.
\end{split}
\end{align}

For $t=t_d+1,\ldots,T$, using smoothness property, we have: 

\begin{align}\nn%
\begin{split}
R_\Sc(\wsbf_{t+1})&\leq R_\Sc(\wsbf_{t})-\alpha_t\nabla R_\Sc(\wsbf_{t})^\top\nabla \loss(\wsbf_t;\zbf_{\isf_t})
+\frac{\beta}{2}\|\nabla \loss(\wsbf_t;\zbf_{\isf_t})\|^2\\
&\leq R_\Sc(\wsbf_{t})-\alpha_t\nabla R_\Sc(\wsbf_{t})^\top\nabla \loss(\wsbf_t;\zbf_{\isf_t})
+\frac{\beta L^2}{2}.
\end{split}
\end{align}
Using a similar argument, we can find the following upper bound when the momentum is set to zero: 

\begin{align}\label{generallrconvKey2}
\begin{split}
\E_{A}\big[\sum_{t=t_d+1}^{T}\alpha_t\|\nabla R_\Sc(\wsbf_{t})\|^2\big]&\leq R_\Sc(\wsbf_{0}) - \inf_{\wbf} R_\Sc(\wbf)+\frac{\beta L^2}{2}\sum_{t=t_d+1}^{T} \alpha_t^2.
\end{split}
\end{align}

Adding~\cref{generallrconvKey1,generallrconvKey2}, we have

\begin{align}\nn%
\begin{split}
&\E_{A}\Big[\sum_{t=1}^{t_d}\frac{1-\mu_d^{t_d-t+1}}{1-\mu_d}\alpha_t\|\nabla R_\Sc(\wsbf_{t})\|^2+\sum_{t=t_d+1}^{T}\alpha_t\|\nabla R_\Sc(\wsbf_{t})\|^2\Big]\\&\leq 2\big(R_\Sc(\wsbf_{0}) - \inf_{\wbf} R_\Sc(\wbf)\big)+\frac{\beta L^2}{2(1-\mu_d)^2}\sum_{t=1}^{t_d} \alpha_t^2+ \frac{\beta\mu_d L^2}{(1-\mu_d)^3}  \sum_{t=1}^{t_d-1} \alpha_t^2 +\frac{\beta L^2}{2}\sum_{t=t_d+1}^{T} \alpha_t^2\\
&\leq 2\big(R_\Sc(\wsbf_{0}) - \inf_{\wbf} R_\Sc(\wbf)\big)+
\max\Big\{\frac{\beta L^2}{2(1-\mu_d)^2},\frac{\beta\mu_d L^2}{(1-\mu_d)^3},\frac{\beta L^2}{2} 
\Big\}\sum_{t=1}^{T} \alpha_t^2.
\end{split}
\end{align}
On the other hand, we find a lower bound on the left hand side: 

\begin{align}\nn%
\begin{split}
&\E_{A}\Big[\sum_{t=1}^{t_d}\frac{1-\mu_d^{t_d-t+1}}{1-\mu_d}\alpha_t\|\nabla R_\Sc(\wsbf_{t})\|^2+\sum_{t=t_d+1}^{T}\alpha_t\|\nabla R_\Sc(\wsbf_{t})\|^2\Big]\\
&\geq \min\Big\{\min_{1\leq t\leq t_d}\frac{1-\mu_d^{t_d-t+1}}{1-\mu_d},1\Big\}
\E_{A}\big[\sum_{t=1}^{T}\alpha_t\|\nabla R_\Sc(\wsbf_{t})\|^2\big]\\
&\geq 
 \sum_{t=1}^{T}\alpha_t\min_{1\leq t\leq T}\E_{A}\big[\|\nabla R_\Sc(\wsbf_{t})\|^2\big].
\end{split}
\end{align}
Combining the above upper bound and lower bound, we have 

\begin{align}\label{app:generallrconvRate}
\begin{split}
\min_{1\leq t\leq T}\E_{A}\big[\|\nabla R_\Sc(\wsbf_{t})\|^2\big]
&\leq \frac{2\big(R_\Sc(\wsbf_{0}) - \inf_{\wbf} R_\Sc(\wbf)\big)}{\sum_{t=1}^{T}\alpha_t}\\
&\quad + \frac{
\max\Big\{\frac{\beta L^2}{2(1-\mu_d)^2},\frac{\beta\mu_d L^2}{(1-\mu_d)^3},\frac{\beta L^2}{2} 
\Big\}\sum_{t=1}^{T} \alpha_t^2}{\sum_{t=1}^{T}\alpha_t}
\end{split}
\end{align}
which completes the proof. In particular, by substituting 
step-size $\alpha_t=\alpha_0/t^q$, we achieve convergence with the rate of  $\Oc(T^{q-1})$ for~\ref{earlyupdate} with any $t_d$.

\section{Convergence of~\ref{earlyupdate} with constant step-size}\label{app:thm:earlymnconconv}

\bth\label{thm:earlymnconconv}  Suppose {that  $\loss$ satisfies~\cref{assu:LipSm} and} that  the~\ref{earlyupdate} update  is executed for $T$ steps with constant step-size $\alpha< 2(1-\mu_d)$ and momentum {$\mu_d\in(0,1)$} in the first $t_d$ steps. Then, for any $\Sc$ and $0<t_d\leq T$, we have
\begin{align}\label{convncon1}
\min_{t=0,\ldots,T}\epsilon(t)\leq \frac{W+J_2}{W_1}
\end{align} where $\epsilon(t)\defeq \E_{A}[\|\nabla_\wbf R_\Sc(\wsbf_{t})\|^2]$, $J_2=(t_d+1)\big(\frac{\beta}{2}\big(\frac{\alpha L}{1-\mu_d}\big)^2+\frac{1}{2}\big(\frac{\alpha \beta L\mu_d}{(1-\mu_d)^2}\big)^2\big)+(T-t_d)\frac{\beta}{2}\alpha^2L^2$, $W_1=(t_d+1)\big(\frac{\alpha}{1-\mu_d}-\frac{\alpha^2}{2(1-\mu_d)^2}\big)+(T-t_d)\alpha$, and $W=\E_{A}[R_\Sc(\wsbf_0)-R_\Sc(\wsbf_\Sc^*)]$ with $\wbf_\Sc^*=\arg\min_{\wbf} R_\Sc(\wbf)$.
\eth
 \bpr
We analyze the convergence of~\ref{earlyupdate} for a {smooth Lipschitz loss} function with constant step-size. To facilitate the convergence analysis, for $t\in[t_d]$, we  define $\psbf_t\defeq \frac{\mu_d}{1-\mu_d}(\wsbf_t-\wsbf_{t-1})$ with $\psbf_0=0$. Substituting this into the~\ref{earlyupdate} update, the parameter recursion is given by%
\begin{align}\label{parrecnonc}
\wsbf_{t+1}+\psbf_{t+1}=\wsbf_t+\psbf_t-\frac{\alpha}{1-\mu_d}\nabla_\wbf \loss(\wsbf_t;\zbf_{\isf_t}).
\end{align} 

We also define $\xsbf_t\defeq \wsbf_t+\psbf_t$. Note that for a $\beta$-smooth function $f$ and for all $\ubf,~\vbf\in\Psi$, we have
\begin{align}\label{betaprop}
f(\ubf)\leq f(\vbf)+\nabla f(\vbf)^\top(\ubf-\vbf)+\frac{\beta}{2}\|\ubf-\vbf\|^2.
\end{align} We note that $\xsbf_{t}=\wsbf_{t}$ for $t>t_d$. Let $t\in[t_d]$. Since the empirical risk $R_\Sc$ is a $\beta$-smooth function, we have
\begin{align}\label{convnconine1}
\begin{split}
R_\Sc(\xsbf_{t+1})&\leq R_\Sc(\xsbf_{t})+\nabla_\wbf R_\Sc(\xsbf_{t})^\top(\xsbf_{t+1}-\xsbf_{t})+\frac{\beta\alpha^2}{2(1-\mu_d)^2}\|\nabla_\wbf \loss(\wsbf_t;\zbf_{\isf_t})\|^2\\
&\leq R_\Sc(\xsbf_{t})+\frac{\beta}{2}\Big(\frac{\alpha L}{1-\mu_d}\Big)^2-\frac{\alpha}{1-\mu_d}\nabla_\wbf R_\Sc(\xsbf_{t})^\top\nabla_\wbf \loss(\wsbf_t;\zbf_{\isf_t})
\end{split}
\end{align} where we use the fact that $\|\nabla_\wbf \loss(\wsbf_t;\zbf_{\isf_t})\|\leq L$, due to the $L$-Lipschitz property.

Upon taking the expectation w.r.t. $\isf_t$ in~\cref{convnconine1} and defining $\rsf_t\defeq R_\Sc(\xsbf_{t+1})-R_\Sc(\xsbf_{t})$, we have
\begin{align}\label{convnconine2}
\begin{split}
\E_{\isf_t}[\rsf_t]&\leq -\frac{\alpha}{1-\mu_d}\nabla_\wbf R_\Sc(\xsbf_{t})^\top\nabla_\wbf R_\Sc(\wsbf_{t})+\frac{\beta}{2}\Big(\frac{\alpha L}{1-\mu_d}\Big)^2\\
&=-\frac{\alpha}{1-\mu_d}\big(\nabla_\wbf R_\Sc(\xsbf_{t})-\nabla_\wbf R_\Sc(\wsbf_{t})\big)^\top\nabla_\wbf R_\Sc(\wsbf_{t})-\frac{\alpha}{1-\mu_d}\|\nabla_\wbf R_\Sc(\wsbf_{t})\|^2+\frac{\beta}{2}\Big(\frac{\alpha L}{1-\mu_d}\Big)^2\\
&\leq \frac{1}{2}\|\nabla_\wbf R_\Sc(\xsbf_{t})-\nabla_\wbf R_\Sc(\wsbf_{t})\|^2+\frac{\beta}{2}\Big(\frac{\alpha L}{1-\mu_d}\Big)^2+\Big(\frac{\alpha^2}{2(1-\mu_t)^2}-\frac{\alpha}{1-\mu_d}\Big)\|\nabla_\wbf R_\Sc(\wsbf_{t})\|^2
\end{split}
\end{align} where the last inequality is obtained using $2\ubf^\top\vbf\leq\|\ubf\|^2+\|\vbf\|^2$. For $t>t_d$, we have

\begin{align}\label{convncon_nomom}
\begin{split}
\E_{\isf_t}[\rsf_t]&\leq -\alpha\nabla_\wbf R_\Sc(\wsbf_{t})^\top\nabla_\wbf R_\Sc(\wsbf_{t})+\frac{\beta}{2}(\alpha L)^2\\
&\leq \frac{\beta}{2}({\alpha L})^2-{\alpha}\|\nabla_\wbf R_\Sc(\wsbf_{t})\|^2.
\end{split}
\end{align}

In the following, we obtain an upper bound on $\|\nabla_\wbf R_\Sc(\xsbf_{t})-\nabla_\wbf R_\Sc(\wsbf_{t})\|^2$ in~\cref{convnconine2} for $t\in[t_d]$. 

Since $R_\Sc$ is $\beta$-smooth, we have
\begin{align}\label{normdiffgrads}
\|\nabla_\wbf R_\Sc(\xsbf_{t})-\nabla_\wbf R_\Sc(\wsbf_{t})\|^2\leq \beta^2\|\xsbf_{t}-\wsbf_{t}\|^2.
\end{align} We also note that $\beta^2\|\xsbf_{t}-\wsbf_{t}\|^2=\frac{\beta^2\mu_d^2}{(1-\mu_d)^2}\|\wsbf_{t}-\wsbf_{t-1}\|^2.$

For notational simplicity, we define $\qsbf_t\defeq \frac{1-\mu_d}{\mu_d}\psbf_t$ with $\qsbf_0=0$. Rewriting the~\ref{earlyupdate} update rule, the parameter recursion is given by
\begin{align}\label{normdiffgradsrec}
\qsbf_{t+1}=\mu_d\qsbf_t-\alpha\nabla_\wbf \loss(\wsbf_t;\zbf_{\isf_t}).
\end{align}
Unraveling the recursion~\cref{normdiffgradsrec}, we have
\begin{align}\label{normdiffgradsrec2}
\begin{split}
\qsbf_{t}&=-\alpha\sum_{k=0}^{t-1}\mu_d^{t-1-k}\nabla_\wbf \loss(\wsbf_k;\zbf_{\isf_k})\\
&=-\alpha\sum_{k=0}^{t-1}\mu_d^{k}\nabla_\wbf \loss(\wsbf_{t-1-k};\zbf_{\isf_{t-1-k}}).
\end{split}
\end{align}
We define $\Theta_{t-1}\defeq\sum_{k=0}^{t-1}\mu_d^{k}=\frac{1-\mu_d^{t}}{1-\mu_d}.$
Then we can find an upper bound on $\|\qsbf_{t}\|$ as follows:
\begin{align}\label{normdiffgradsrecineq}
\begin{split}
\|\qsbf_{t}\|&=\|-\alpha\sum_{k=0}^{t-1}\mu_d^{k}\nabla_\wbf \loss(\wsbf_{t-1-k};\zbf_{\isf_{t-1-k}})\|\\
&=\alpha\|\sum_{k=0}^{t-1}\mu_d^{k}\nabla_\wbf \loss(\wsbf_{t-1-k};\zbf_{\isf_{t-1-k}})\|\\
&\leq \alpha\sum_{k=0}^{t-1}\mu_d^{k}\|\nabla_\wbf \loss(\wsbf_{t-1-k};\zbf_{\isf_{t-1-k}})\|\\
&\leq \alpha\Theta_{t-1} L\\
&\leq \frac{\alpha L}{1-\mu_d}.
\end{split}
\end{align} Substituting~\cref{normdiffgradsrecineq} into~\cref{normdiffgrads},  we obtain the following upper bound on $\|\nabla_\wbf R_\Sc(\xsbf_{t})-\nabla_\wbf R_\Sc(\wsbf_{t})\|^2$:
\begin{align}\label{normdiffgradsineq}
\!\!\!\!\!\|\nabla_\wbf R_\Sc(\xsbf_{t})-\nabla_\wbf R_\Sc(\wsbf_{t})\|^2\leq \frac{\alpha^2 \beta^2L^2\mu_d^2}{(1-\mu_d)^4}.
\end{align}

Substituting~\cref{normdiffgradsineq} into \cref{convnconine2} and taking expectation over $\isf_0,\ldots,\isf_{t}$, we have
\begin{align}\label{convnconine3}
\E_{A}[\rsf_t]&\leq -\Big(\frac{\alpha}{1-\mu_d}-\frac{\alpha^2}{2(1-\mu_d)^2}\Big)\E_{A}[\|\nabla_\wbf R_\Sc(\wsbf_{t})\|^2]+\frac{\beta}{2}\Big(\frac{\alpha L}{1-\mu_d}\Big)^2+\frac{1}{2}\Big(\frac{\alpha \beta L\mu_d}{(1-\mu_d)^2
}\Big)^2.
\end{align}

Summing~\cref{convnconine3} for $t\in[t_d]$ and~\cref{convncon_nomom} for $t=t_d+1,\ldots,T$, we have
\begin{align}\label{convnconine4}
\begin{split}
J_1&\leq \E_{A}[R_\Sc(\xsbf_{0})-R_\Sc(\xsbf_{t+1})]+J_2\\
&\leq \E_{A}[R_\Sc(\wsbf_{0})-R_\Sc(\wsbf_{\Sc}^*)]+J_2
\end{split}
\end{align} where
\begin{align}\label{J1}J_1&=\Big(\frac{\alpha}{1-\mu_d}-\frac{\alpha^2}{2(1-\mu_d)^2}\Big)\sum_{t=0}^{t_d}\E_{A}[\|\nabla_\wbf R_\Sc(\wsbf_{t})\|^2]+\alpha\sum_{t=t_d+1}^T\E_{A}[\|\nabla_\wbf R_\Sc(\wsbf_{t})\|^2]
\end{align} and

\begin{align}\nn
J_2=(t_d+1)\Big(\frac{\beta}{2}\Big(\frac{\alpha L}{1-\mu_d}\Big)^2+\frac{1}{2}\Big(\frac{\alpha \beta L\mu_d}{(1-\mu_d)^2}\Big)^2\Big)+\big(T-t_d\big)\frac{\beta}{2}\alpha^2L^2.
\end{align}

Noting $\alpha\leq 2(1-c)(1-\mu_d)$ for some $0<c<1$, we obtain the following lower bound on $J_1$ in~\cref{J1}:
\begin{align}\label{J1ineq}
\begin{split}
J_1&\geq \frac{(t_d+1)\alpha c}{1-\mu_d}\min_{t=0,\ldots,t_d}\epsilon(t)+(T-t_d)\alpha\min_{t=t_d+1,\ldots,T}\epsilon(t)\\
&\geq\chi_3\min_{t=0,\ldots,T}\epsilon(t)
\end{split}
\end{align} where $\chi_3= \frac{(t_d+1)\alpha c}{1-\mu_d}+(T-t_d)\alpha$.

Substituting~\cref{J1ineq} into \cref{convnconine4}, we obtain \cref{convncon1}, which completes the proof.

\epr
{We cannot directly combine standard bounds for~\ref{update} and SGD to analyze convergence of~\ref{earlyupdate}  because the standard analysis requires characterization of the empirical risk at $\wsbf_{t_d}$.  Instead our proof is inspired by the convergence proof for~\ref{update} by carefully handling time-varying momentum.}
We now study the upper bound \eqref{convncon1} as a function of $t_d$ for a given $\mu_d$. Note that the first term in the upper bound vanishes as $T\rightarrow \infty$. 

\begin{remark} In~\cref{app:cor:SGDEMvstd}, we provide a sufficient condition for the upper bound \eqref{convncon1} to become a monotonically decreasing function of $t_d$.  In~\cref{thm:earlymnconconv}, $\frac{J_2}{W_1}\approx\frac{aT+bt_d}{cT+dt_d}$ for some $a,b,c,d$. We may provide a looser bound by establishing an upper bound on $J_2$ and a lower bound on $W_1$. However,  such looser bound is not useful since we will not be able to recover standard bounds for SGD and~\ref{update}. In order to provide a simpler expression
and understand how adding momentum affects the convergence, in~\cref{app:momentumrole} we study the convergence bound for a special form of~\ref{earlyupdate} and show the benefit of using momentum. We also provide a simple sufficient condition for the non-vanishing term in the convergence bound to become a monotonically decreasing function of $\mu_d$.
\end{remark}

We also establish convergence guarantees for~\ref{earlyupdate} with another time-dependent step-size in~\cref{app:timelr_decay}.

\begin{remark} {In~\cref{thm:earlymnconconv}, our focus is on guaranteeing convergence to a local minimum, which holds for {\it any} $t_d\leq T$.  We note that optimizing the upper bound  in \eqref{convncon1}}  over $t_d$  will not provide much intuition on the optimal $t_d$ in terms of training error since we cannot guarantee the actual suboptimality gap (optimization error) of nonconvex loss functions. In practice, we need to tune $t_d$ when training, \eg neural networks. Our experimental results show that a nontrivial $t_d$ can be optimal in terms of test error.
\end{remark}

\section{Convergence guarantees for~\ref{earlyupdate} with time-dependent and time-decaying step-sizes}\label{app:timelr_decay}
We establish convergence guarantees for~\ref{earlyupdate} with time-dependent and time-decaying step-sizes as follows.  
 \bth\label{thm:earlymnconconv_timelr} Suppose that  $\loss$ satisfies~\cref{assu:LipSm} and that the~\ref{earlyupdate} update is executed for $T$ steps with momentum $\mu_d$ in the first $t_d$ steps and time-dependent step-size $\alpha=\min\{2(1-c)(1-\mu_d),\frac{K}{\max\{\sqrt{t_d+1},\sqrt{T-t_d}\}}\}$ for some $0<c<1$ and $0<K$. Then, for any $\Sc$ and $0<t_d\leq T$, we have
\begin{align}\label{convncon_timelr}
\min_{t=0,\ldots,T}\epsilon(t)\leq\frac{\tilde T(W+\tilde J_2)}{\tilde W_1}
\end{align} where
\begin{align}
\tilde J_2&=\frac{\beta}{2}\big(\frac{K L}{1-\mu_d}\big)^2+\frac{1}{2}\big(\frac{K \beta L\mu_d}{(1-\mu_d)^2}\big)^2+\frac{\beta}{2}K^2L^2,\nn\\
\tilde W_1&=\frac{(t_d+1)c}{1-\mu_d}+T-t_d,\nn\\
\tilde T &= \max\{\frac{1}{2(1-c)(1-\mu_d)},\frac{\max\{\sqrt{t_d+1},\sqrt{T-t_d}\}}{K}\}.\nn
\end{align}
\eth

\bth\label{thm:earlymnconconv_decay_app}Suppose that  $\loss$ satisfies~\cref{assu:LipSm} and that  the~\ref{earlyupdate} update is executed for $T$ steps with time-decaying step-size $\alpha_t=\frac{\alpha_0}{t+1}$ for $t=0,1,\ldots,T$ with $\alpha_0\leq 2(1-c)(1-\mu_d)$ for some $0<c<1$ and momentum $\mu_d>\exp(-1)$ in the first $t_d$ steps. Then, for any $\Sc$ and $0<t_d\leq T$, we have
\begin{align}\label{convncon1_decay_app}
\min_{t=0,\ldots,T}\epsilon(t)\leq \frac{W+\hat J_2}{\hat W_1}
\end{align} where  
\begin{align}
\hat J_2&=\beta(\frac{\alpha_0 L}{1-\mu_d})^2+\frac{\beta}{2}(\alpha_0 L)^2 \frac{1}{t_d+1}+\sum_{t=1}^{t_d}\big(\frac{\alpha_0\overline c_t \beta L\mu_d}{1-\mu_d}\big)^2,\nn\\
\hat W_1&=\frac{\ln(t_d+1)\alpha_0 c}{1-\mu_d}+\ln\big(\frac{T}{t_d+2}\big)\alpha_0,\nn\\
\overline c_t &= \min\{\frac{1}{1-\mu_d}, 1+\ln(t),\mu_d^{t}(\mu_d^{-1}+I(t))\},~\and~I(t) = \int_1^t \frac{\mu_d^{-u}}{u}\ud u.\nn
\end{align}
\eth
\bpr  Following the proof of~\cref{thm:earlymnconconv} for $t\in[t_d]$, we have
\begin{align}\label{decay_convnconine2}
\E_{\isf_t}[\rsf_t]&\leq \frac{1}{2}\|\nabla_\wbf R_\Sc(\xsbf_{t})-\nabla_\wbf R_\Sc(\wsbf_{t})\|^2+\frac{\beta}{2}\Big(\frac{\alpha_t L}{1-\mu_d}\Big)^2+\Big(\frac{\alpha_t^2}{2(1-\mu_d)^2}-\frac{\alpha_t}{1-\mu_d}\Big)\|\nabla_\wbf R_\Sc(\wsbf_{t})\|^2.
\end{align} 

For $t>t_d$, we have $\xsbf_{t}=\wsbf_{t}$ and

\begin{align}\label{decay_convncon_nomom}
\begin{split}
\E_{\isf_t}[\rsf_t]&\leq -\alpha_t\nabla_\wbf R_\Sc(\wsbf_{t})^\top\nabla_\wbf R_\Sc(\wsbf_{t})+\frac{\beta}{2}(\alpha_t L)^2\\
&\leq \frac{\beta}{2}({\alpha_t L})^2-{\alpha_t}\|\nabla_\wbf R_\Sc(\wsbf_{t})\|^2.
\end{split}
\end{align}

In the following, we obtain an upper bound on $\|\nabla_\wbf R_\Sc(\xsbf_{t})-\nabla_\wbf R_\Sc(\wsbf_{t})\|^2$ in~\cref{decay_convnconine2}. %
Since $R_\Sc$ is $\beta$-smooth, we have
\begin{align}\label{decay_normdiffgrads}
\|\nabla_\wbf R_\Sc(\xsbf_{t})-\nabla_\wbf R_\Sc(\wsbf_{t})\|^2\leq \beta^2\|\xsbf_{t}-\wsbf_{t}\|^2.
\end{align} We also note that 
\begin{align}\nn
\beta^2\|\xsbf_{t}-\wsbf_{t}\|^2=\frac{\beta^2\mu_d^2}{(1-\mu_d)^2}\|\wsbf_{t}-\wsbf_{t-1}\|^2.
\end{align}

For notational simplicity, we define $\qsbf_t\defeq \frac{1-\mu_d}{\mu_d}\psbf_t$ with $\qsbf_0=0$. Rewriting the~\eqref{earlyupdate} update rule, the parameter recursion is given by
\begin{align}\label{decay_normdiffgradsrec}
\qsbf_{t+1}=\mu_d\qsbf_t-\alpha_t\nabla_\wbf \loss(\wsbf_t;\zbf_{\isf_t}).
\end{align}
Unraveling the recursion~\cref{decay_normdiffgradsrec}, we have
\begin{align}\label{decay_normdiffgradsrec2}
\qsbf_{t}&=-\alpha_0\sum_{k=0}^{t-1}\frac{\mu_d^{t-1-k}}{k+1}\nabla_\wbf \loss(\wsbf_k;\zbf_{\isf_k}).
\end{align}

\blm Provided that $\mu_d\geq \exp(-1)$, we have $\|\qsbf_{t}\|^2\leq \overline c_t^2\alpha_0^2L^2$ for $t\leq t_d$, where
\begin{align}\nn
\overline c_t = \min\{\frac{1}{1-\mu_d}, 1+\ln(t),\mu_d^{t}(\mu_d^{-1}+I(t))\}~\and~I(t) = \int_1^t \frac{\mu_d^{-u}}{u}\ud u.
\end{align} 
\bpr
Following the proof of~\cref{thm:earlymnconconv}, an upper bound on $\|\qsbf_{t}\|^2$ is given by 
\begin{align}\nn
\|\qsbf_{t}\|^2 \leq \alpha_0^2L^2\tilde S^2
\end{align}
where
\begin{align}\nn
\tilde S  = \sum_{k=0}^{t-1}\frac{\mu_d^{t-1-k}}{k+1}.
\end{align}
  Note that %
\begin{align}\nn
\tilde S\leq \sum_{k=0}^{t-1}\mu_d^{k}=\frac{1-\mu_d^{t}}{1-\mu_d}~\and~\tilde S\leq \sum_{k=1}^{t}1/k\leq 1+\int_1^t 1/u\ud u=  1+\ln(t).
\end{align}

 Rewriting $\tilde S$ as $\tilde S = \mu_d^{t}\sum_{k=1}^t\frac{\mu_d^{-k}}{k}$ and noting $f(u) = \mu_d^{-u}/{u}$ is convex and non-increasing for $1\leq u\leq t$ due to the lower bound $\mu_d\geq \exp(-1)$. Therefore, we have $\tilde S\leq\mu_d^{t}(\mu_d^{-1}+I(t))$. 
\epr
\elm

Substituting the upper bound on $\|\qsbf_{t}\|^2$ into~\cref{decay_normdiffgrads},  we obtain the following upper bound on $\|\nabla_\wbf R_\Sc(\xsbf_{t})-\nabla_\wbf R_\Sc(\wsbf_{t})\|^2$:
\begin{align}\label{decay_normdiffgradsineq}
\!\!\!\!\!\|\nabla_\wbf R_\Sc(\xsbf_{t})-\nabla_\wbf R_\Sc(\wsbf_{t})\|^2\leq \frac{\alpha_0^2\overline c_t^2 \beta^2L^2\mu_d^2}{(1-\mu_d)^2}.
\end{align}

Substituting~\cref{decay_normdiffgradsineq} into \cref{decay_convnconine2} and taking expectation over $\isf_0,\ldots,\isf_{t}$, we have
\begin{align}\label{decay_convnconine3}
\E_{A}[\rsf_t]&\leq -\Big(\frac{\alpha_t}{1-\mu_t}-\frac{\alpha_t^2}{2(1-\mu_d)^2}\Big)\E_{A}[\|\nabla_\wbf R_\Sc(\wsbf_{t})\|^2]+\frac{\beta}{2}\Big(\frac{\alpha_t L}{1-\mu_d}\Big)^2+\Big(\frac{\alpha_0\overline c_t \beta L\mu_d}{1-\mu_d}\Big)^2.
\end{align}

Summing~\cref{decay_convnconine3} for $t\in[t_d]$ and~\cref{decay_convncon_nomom} for $t=t_d+1,\ldots,T$, we have
\begin{align}\label{decay_convnconine4}
\begin{split}
\hat J_1&\leq \E_{A}[R_\Sc(\xsbf_{0})-R_\Sc(\xsbf_{t+1})]+\hat J_2\\
&\leq \E_{A}[R_\Sc(\wsbf_{0})-R_\Sc(\wsbf_{\Sc}^*)]+\hat J_2
\end{split}
\end{align} where
\begin{align}\label{decay_J1}\hat J_1&=\sum_{t=0}^{t_d}\Big(\frac{\alpha_t}{1-\mu_d}-\frac{\alpha_t^2}{2(1-\mu_d)^2}\Big)\E_{A}[\|\nabla_\wbf R_\Sc(\wsbf_{t})\|^2]+\sum_{t=t_d+1}^T\alpha_t\E_{A}[\|\nabla_\wbf R_\Sc(\wsbf_{t})\|^2].
\end{align}

Noting $\alpha_0\leq 2(1-c)(1-\mu_d)$ for some $0<c<1$, we obtain the following lower bound on $\hat J_1$ in~\cref{decay_J1}:
\begin{align}\label{decay_J1ineq}
\begin{split}
\hat J_1&\geq \sum_{t=0}^{t_d}\frac{\alpha_t c}{1-\mu_d}\min_{t=0,\ldots,t_d}\epsilon(t)+\sum_{t=t_d+1}^{T}\alpha_t\min_{t=t_d+1,\cdots,T}\epsilon(t)\\
&\geq\hat \chi_3\min_{t=0,\ldots,T}\epsilon(t)
\end{split}
\end{align} where 
\begin{align}
\hat \chi_3=  \frac{\ln(t_d+1)\alpha_0 c}{1-\mu_d}+\ln\big(\frac{T}{t_d+2}\big)\alpha_0.
\end{align}

Finally, we note that 
\begin{align}
\frac{\beta}{2}\Big(\frac{\alpha_0 L}{1-\mu_d}\Big)^2\sum_{t=0}^{t_d}\frac{1}{(t+1)^2}\leq \beta\Big(\frac{\alpha_0 L}{1-\mu_d}\Big)^2\quad\and\quad\frac{\beta}{2}(\alpha_0 L)^2\sum_{t=t_d+1}^{T}\frac{1}{(t+1)^2}\leq \frac{\beta}{2}(\alpha_0 L)^2 \frac{1}{t_d+1}.\nn 
\end{align}

\epr

\section{Sufficient condition for the upper bound in~\cref{thm:earlymnconconv}}\label{app:cor:SGDEMvstd}
In the following corollary, we provide a simple sufficient condition for the upper bound \eqref{convncon1} to become a monotonically decreasing function of $t_d$.

\bcr\label[corollary]{cor:SGDEMvstd} Suppose that  $\loss$ satisfies~\cref{assu:LipSm} and that~\ref{earlyupdate} is executed for {finite} $T$ steps with constant step-size $\alpha< 2c(1-\mu_d)$ with some $c<\frac{1}{2-\mu_d}$ and momentum $\mu_d$ in the first $t_d$ steps. Then the upper bound \eqref{convncon1} is a monotonically decreasing function of $t_d$ if the following condition is satisfied:
\begin{align}
W>\frac{(K_1-K_2)(K_3+TK_4)}{K_3-K_4}-K_1-TK_2
\end{align} where $K_1=\frac{\beta}{2}\big(\frac{\alpha L}{1-\mu_d}\big)^2+\frac{1}{2}\big(\frac{\alpha \beta L\mu_d}{(1-\mu_d)^2}\big)^2$, $K_2=\frac{\beta}{2}\alpha^2L^2$, $K_3=\frac{\alpha}{1-\mu_d}-\frac{\alpha^2}{2(1-\mu_d)^2}$, and $K_4=\alpha$.
\ecr

\bpr Note that we can express the upper bound \eqref{convncon1} as 
\begin{align}\nn
U(t_d)=\frac{W+K_1+TK_2+t_d(K_1-K_2)}{K_3+TK_4+t_d(K_3-K_4)}.
\end{align} 
The proof follows by taking the first derivative of $U$ w.r.t. $t_d$.
\epr 

\cref{cor:SGDEMvstd} implies that adding momentum for a longer time is particularly useful when our initial parameter is sufficiently far from a local minimum.

\section{Understanding the role of momentum on convergence}\label{app:momentumrole} 
In order to understand how adding momentum affects the convergence, we study the convergence bound for a special form of~\ref{earlyupdate} and show the benefit of using momentum. 
\bcr
Suppose we set $t_d=T$ with constant step-size $\alpha< 2(1-\mu_d)$. Then, for any $\Sc$, we have

\begin{align}\label{convnconcor1}
\min_{t=0,\cdots,T}\epsilon(t)\leq \frac{W}{(T+1)\big(\frac{\alpha}{1-\mu_d}-\frac{\alpha^2}{2(1-\mu_d)^2}\big)}+\frac{\beta\alpha^2L^2+\frac{(\alpha \beta L\mu_d)^2}{(1-\mu_d)^2}}{2\alpha(1-\mu_d)-\alpha^2}.
\end{align}
\ecr

Note that the upper bound \eqref{convnconcor1} is a function of $\mu_d$. The first term in the upper bound vanishes as $T\rightarrow \infty$. In the following corollary, we provide a simple sufficient condition for the non-vanishing term in the upper bound \eqref{convnconcor1} to become a monotonically decreasing function of $\mu_d$.
\bcr\label[corollary]{cr:mu_d} Suppose $\beta\leq \frac{2\mu_d-\mu_d^2}{(1-\mu_d)^2}$ and we set $t_d=T$ with $\alpha\leq 2c(1-\mu_d)$ for some $0<c<1$. Then the non-vanishing term in the upper bound \eqref{convncon1} is a monotonically decreasing function of $\mu_d$.
\ecr
\bpr
Noting $\alpha\leq 2c(1-\mu_d)$ for some $0<c<1$, we obtain the following lower bound on $J_1$ in \cref{J1}:
\begin{align}\label{J1ineq2}
J_1\geq\chi_4\min_{t=0,\ldots,T}\epsilon(t)
\end{align} where 
\begin{equation}\nn
\chi_4\defeq\frac{(t_d+1)\alpha (1-c)}{1-\mu_d}+T-t_d.
\end{equation}

Substituting~\cref{J1ineq2} into~\cref{convncon1}, the non-vanishing term in the upper bound $\frac{2L^2\beta(1-\mu_d)}{\alpha (1-c)}c^2\Big(1+\beta\frac{\mu_d^2}{(1-\mu_d)^2}\Big)$ becomes a function of $\mu_d$ through $(1-\mu_d)\Big(1+\beta\frac{\mu_d^2}{(1-\mu_d)^2}\Big)$. We can prove the proposition by taking the first derivative w.r.t. $\mu_d$.%
\epr

\section{High-probability generalization bounds}\label{app:highprob}

We consider the generalization error that depends on a random set of $n$ samples $\Sc$ drawn i.i.d. from some space $\Zc$ with an unknown distribution $D$: 
\begin{align}\nn
\epsilon_g(\Sc)= \E_{A}[R(\wsbf_T)-R_\Sc(\wsbf_T)].
\end{align}
We establish high-probability bounds for generalization error of~\ref{earlyupdate} along the lines of~\citep{feldman2018generalization}.
\bth[High-probability generalization bound]
Let $0<\delta<1$. For the setting described in~\cref{cr:earlymstab1}, with probability at least $1-\delta$ over $\Sc\sim D^n$, the generalization error of~\ref{earlyupdate} $\epsilon_g(\Sc)$ is bounded by $\Oc\Big(\sqrt{\big(\frac{\exp(\mu_d)T^u}{n}+\frac{1}{n}\big)\log\big(\frac{1}{\delta}\big)}\Big)$. 

\bpr The proof follows by the arguments in~\citep[Theorem 1.2]{feldman2018generalization} and substituting the stability upper bound in~\cref{cr:earlymstab1}.
\epr
\eth

\section{Proof of~\cref{thm:sconstab}}\label{app:thm:sconstab}
We track the divergence of two different iterative sequences of update rules with the same starting point.
We remark that our analysis is more involved than \citep{Hardt} as the presence of momentum term requires a more careful bound on the iterative expressions.

To keep the notation uncluttered, we first consider~\ref{update} without projection and defer the discussion of projection to the end of this proof.
Let $\Sc=\{\zbf_1,\ldots,\zbf_n\}$ and $\Sc'=\{\zbf'_1,\ldots,\zbf'_n\}$ be two samples of size $n$ that differ in at most one example. Let $\wsbf_T$ and $\wsbf'_T$ denote the outputs of SGDM on $\Sc$ and $\Sc'$, respectively. We consider the updates $\wsbf_{t+1}=G_t(\wsbf_t)+\mu(\wsbf_t-\wsbf_{t-1})$ and $\wsbf'_{t+1}=G'_t(\wsbf'_t)+\mu(\wsbf'_t-\wsbf'_{t-1})$ with $G_t(\wsbf_t)=\wsbf_t-\alpha\nabla_{\wbf}\loss(\wsbf_t;\zbf_{\isf_t})$ and $G'_t(\wsbf'_t)=\wsbf'_t-\alpha\nabla_{\wbf}\loss(\wsbf'_t;\zbf'_{\isf_t})$, respectively, for $t=1,\cdots,T$. We denote $\delta_t\defeq \|\wsbf_t-\wsbf'_t\|$. Suppose $\wsbf_0=\wsbf'_0$, \ie $\delta_0=0$.

We first establish an upper bound on $\E_A[\delta_T]$.
At step $t$, with probability $1-1/n$, the example is the same in both $\Sc$ and $\Sc'$, \ie $\zbf_{\isf_t}=\zbf'_{\isf_t}$, which implies $G_t=G'_t$. Then $G_t$ becomes $\big(1-\frac{\alpha\beta\gamma}{\beta+\gamma}\big)$-expansive for $\alpha\leq\frac{2}{\beta+\gamma}$ (see, \eg \citep[Appendix A]{Hardt}). Hence, we have
\begin{align}\label{scstineq1}
\begin{split}
\delta_{t+1}&=\|\mu(\wsbf_t-\wsbf'_t)-\mu(\wsbf_{t-1}-\wsbf'_{t-1})+G_t(\wsbf_t)-G_t(\wsbf'_t)\|\\
&\leq \mu\|\wsbf_t-\wsbf'_t\|+\mu\|\wsbf_{t-1}-\wsbf'_{t-1}\|+\|G_t(\wsbf_t)-G_t(\wsbf'_t)\|\\
&\leq \vartheta\delta_t+\mu\delta_{t-1} 
\end{split}
\end{align} where $\vartheta=1+\mu-\frac{\alpha\beta\gamma}{\beta+\gamma}$. With probability $1/n$, the selected example is different in $\Sc$ and $\Sc'$. In this case, we have
\begin{align}\label{scstineq2}
\begin{split}
\delta_{t+1}&=\|\mu(\wsbf_t-\wsbf'_t)-\mu(\wsbf_{t-1}-\wsbf'_{t-1})+G_t(\wsbf_t)-G'_t(\wsbf'_t)\|\\
&\leq \mu\|\wsbf_t-\wsbf'_t\|+\mu\|\wsbf_{t-1}-\wsbf'_{t-1}\|+\phi_3\\
&\leq \vartheta\delta_t+\mu\delta_{t-1}+\|G_t(\wsbf'_t)-G'_t(\wsbf'_t)\|\\
&\leq \vartheta\delta_t+\mu\delta_{t-1}+\|\wsbf'_t-G_t(\wsbf'_t)\|+\|\wsbf'_t-G'_t(\wsbf'_t)\|\\
&\leq \vartheta\delta_t+\mu\delta_{t-1}+2\alpha L
\end{split}
\end{align} where $\phi_3=\|G_t(\wsbf_t)+G_t(\wsbf'_t)-G_t(\wsbf'_t)-G'_t(\wsbf'_t)\|$. The last inequality in \eqref{scstineq2} holds due to the $L$-Lipschitz property. Combining~\cref{scstineq1,scstineq2}, we have
\begin{align}
\begin{split}
\E_A[\delta_{t+1}]&\leq (1-1/n)\big(\vartheta\E_A[\delta_t]+\mu\E_A[\delta_{t-1}]\big)+1/n\big(\vartheta\E_A[\delta_t]+\mu\E_A[\delta_{t-1}]+2\alpha L\big)\\
&=\vartheta\E_A[\delta_t]+\mu\E_A[\delta_{t-1}]+\frac{2\alpha L}{n}.
\end{split}
\end{align}
Let us consider the recursion
\begin{align}
\E_A[\tilde\delta_{t+1}]=\vartheta\E_A[\tilde\delta_t]+\mu\E_A[\tilde\delta_{t-1}]+\frac{2\alpha L}{n}\label{eq:rec1}
\end{align} with $\tilde\delta_0=\delta_0=0$.  Upon inspecting~\cref{eq:rec1} it is clear that \begin{align}\E_A[\tilde\delta_{t}]\geq\vartheta\E_A[\tilde\delta_{t-1}], \qquad \forall t \ge 1,\label{eq:rec2}\end{align}
as we simply drop the remainder of positive terms. Substituting~\cref{eq:rec2} into~\cref{eq:rec1}, we have
\begin{align}
\begin{split}
\E_A[\tilde\delta_{t+1}]&\leq\Big(1+\mu+\frac{\mu}{\vartheta}-\frac{\alpha\beta\gamma}{\beta+\gamma}\Big)\E_A[\tilde\delta_t]+\frac{2\alpha L}{n}\\
&\leq\big(\vartheta+2\mu\big)\E_A[\tilde\delta_t]+\frac{2\alpha L}{n}
\end{split}
\end{align} where the second inequality holds due to $\mu\geq \frac{\alpha\beta\gamma}{\beta+\gamma}-\frac{1}{2}$.

Noting that $\E_A[\tilde\delta_{t}]\geq\E_A[\delta_{t}]$ for all $t$ including $T$, we have
\begin{align}\nn
\E_A[\delta_T]\leq\frac{2\alpha L}{n}\sum_{t=1}^T(\vartheta+2\mu)^t \le \frac{2\alpha L(\beta+\gamma)}{n\big(\alpha\beta\gamma-3\mu(\beta+\gamma)\big)}
\end{align}  where the second expression holds since $0\le\mu<\frac{\alpha\beta\gamma}{3(\beta+\gamma)}$.

Applying the $L$-Lipschitz property on $\loss(\cdot,\zbf)$, it follows
\begin{align}
\E_A[|\loss(\wsbf_T;\zbf)-\loss(\wsbf'_T;\zbf)|]&\leq L\E_A[\delta_T]\nn\\
&\leq \frac{2\alpha L^2(\beta+\gamma)}{n\big(\alpha\beta\gamma-3\mu(\beta+\gamma)\big)}.
\end{align}
Since this bound holds for all $\Sc$, $\Sc'$, and $\zbf$, we obtain an upper bound on the uniform stability and the proof is complete.

Our stability bound in above holds for the~\eqref{projupdate} update  because Euclidean projection onto a convex set does not increase the distance between projected points \citep{Rockafellar}. In particular, note that inequalities \eqref{scstineq1} and \eqref{scstineq2} still hold under~\ref{projupdate}.

\section{Convergence bound for strongly convex loss}\label{app:thm:sconopt}
In this section, we develop an upper bound on the optimization error for the case of strongly convex loss,  which is defined as
\begin{align}\label{opterr}
\epsilon_{\opt}\defeq \E_{\Sc,A}[R_\Sc(\hat\wbf_T)-R_\Sc(\wbf_\Sc^*)]
\end{align} where $\hat\wsbf_T$ denotes the average of $T$ steps of the algorithm, \ie $\hat\wsbf_T=\frac{1}{T+1}\sum_{t=0}^T\wsbf_t$, $R_\Sc(\wbf)= \frac{1}{n}\sum_{i=1}^n\loss(\wbf;\zbf_i)$, and $\wbf_\Sc^*=\arg\min_{\wbf} R_\Sc(\wbf)$.

The optimization error quantifies the gap between the empirical risk of~\ref{projupdate} and the optimal empirical risk.

\bth\label{thm:sconopt}
Suppose that  $\loss$ satisfies~\cref{assu:LipSm,assu:SC} and that~\ref{projupdate} is executed   for $T$ steps with constant step-size $\alpha$ and momentum $\mu$. Then we have\footnote{Linear convergence results for SGD can be obtained under a stringent condition \citep{Needell}. Such a condition requires that the loss function is simultaneously minimized on each training example, and it does not apply to our setting. Different from \citep{Yang,Ghadimi}, we analyze the convergence of~\ref{projupdate} for a smooth and strongly convex loss function with constant step-size.}
\begin{align}\label{conv}
\epsilon_{\opt}\leq \frac{\mu W_0}{(1-\mu)T}+\frac{(1-\mu)W_1}{2\alpha T}-\frac{\gamma W_2}{2}-\frac{\mu\gamma W_3}{2(1-\mu)}+\frac{\alpha L^2}{2(1-\mu)}
\end{align} where $W_0=\E_{\Sc,A}[R_\Sc(\wsbf_0)-R_\Sc(\wsbf_T)]$, $W_1=\E_{\Sc,A}[\|\wsbf_0-\wsbf_\Sc^*\|^2]$, $W_2=\E_{\Sc,A}[\|\hat\wsbf_T-\wsbf_\Sc^*\|^2]$, and $W_3=\frac{1}{T+1}\sum_{t=0}^T\E_{\Sc,A}[\|\wsbf_t-\wsbf_{t-1}\|^2]$.%
\eth
\bpr
Again, we first consider~\ref{update} without projection and discuss the extension to projection at the end of this proof.
To facilitate the convergence analysis, we  define:
\begin{align}\label{vectorp}
\psbf_t\defeq \frac{\mu}{1-\mu}(\wsbf_t-\wsbf_{t-1})
\end{align} with $\psbf_0=0$. Substituting into~\ref{update}, we have 
\begin{align}\label{parrec2}
\|\ssbf_{t+1}-\wbf\|^2=&\|\ssbf_t-\wbf\|^2+\big(\frac{\alpha}{1-\mu}\big)^2\|\nabla_\wbf \loss(\wsbf_t;\zbf_{\isf_t})\|^2-\frac{2\alpha}{1-\mu}(\ssbf_t-\wbf)^\top\nabla_\wbf \loss(\wsbf_t;\zbf_{\isf_t})
\end{align} where $\ssbf_t=\wsbf_t+\psbf_t$. Substituting $\ssbf_t$, taking the expectation w.r.t. $\isf_t$, using the $L$-Lipschitz assumption, noting $R_\Sc$ is a $\gamma$-strongly convex function, summing for $t=0,\ldots,T$, and rearranging terms, we have 

\begin{align}\label{recb}\textstyle
\begin{split}
\varsigma_T&\leq\frac{2\alpha\mu}{(1-\mu)^2} \E_A[R_\Sc(\wsbf_0)-R_\Sc(\wsbf_T)]-\frac{\alpha\gamma}{1-\mu}\sum_{t=0}^T\E_A[\|\wsbf_t-\wbf\|^2]\\
&\quad+ \frac{\alpha^2 L^2(T+1)}{(1-\mu)^2}+\E_A[\|\wsbf_0-\wbf\|^2]-\frac{\alpha\mu\gamma}{(1-\mu)^2}\sum_{t=0}^T \E_A[\|\wsbf_t-\wsbf_{t-1}\|^2]
\end{split}
\end{align} where $\varsigma_T=\frac{2\alpha}{1-\mu}\sum_{t=0}^T\E_A[R_\Sc(\wsbf_t)-R_\Sc(\wbf)]$.
Since $\|\cdot\|^2$ is a convex function, we have $\|\hat\wbf_T-\wbf\|^2\leq\frac{1}{T+1}\sum_{t=0}^T\|\wbf_t-\wbf\|^2$ for all $\wbf_T$ and $\wbf$. Furthermore, due to the convexity of $R_S$, we have
\begin{align}
R_\Sc(\hat\wbf_T)-R_\Sc(\wbf)\leq \frac{1}{T+1}\sum_{t=0}^T\big(R_\Sc(\wbf_t)-R_\Sc(\wbf)\big).
\end{align}
 
Taking expectation over $\Sc$, applying the above inequalities, and substituting $\wbf=\wsbf_\Sc^*$, we obtain~\cref{conv}.

Our convergence bound in \eqref{conv} can be extended to~\ref{projupdate}. Let use denote
\begin{align}\nn
\ybf_{t+1}\defeq \wsbf_t+\mu(\wsbf_t-\wsbf_{t-1})-\alpha\nabla_\wbf \loss(\wsbf_t;\zbf_{\isf_t}).
\end{align}

Then, for any feasible $\wbf\in\Omega$,~\cref{parrec2} holds for $\ysbf_{t+1}$, \ie
\begin{align}\label{projparrec2}
\begin{split}
\|\hat\ysbf_{t+1}-\wbf\|^2&=\|\ssbf_t-\wbf\|^2+\big(\frac{\alpha}{1-\mu}\big)^2\|\nabla_\wbf \loss(\wsbf_t;\zbf_{\isf_t})\|^2\\
&\quad-\frac{2\alpha}{1-\mu}(\ssbf_t-\wbf)^\top\nabla_\wbf \loss(\wsbf_t;\zbf_{\isf_t})
\end{split}
\end{align} where $\hat\ysbf_t=\ysbf_t+\frac{\mu}{1-\mu}(\ysbf_t-\wsbf_{t-1})$. 

Note that the LHS of \cref{projparrec2} can be written as
\begin{align}\nn\textstyle
\|\hat\ysbf_{t+1}-\wbf\|^2=\frac{1}{(1-\mu)^2}\|\ysbf_{t+1}-\big(\mu\wsbf_t+(1-\mu)\wbf\big)\|^2.
\end{align} We note that $\tilde\wsbf_t=\mu\wsbf_t+(1-\mu)\wbf\in\Omega$ for any $\wbf\in\Omega$ and $\wsbf_t\in\Omega$ since $\Omega$ is convex.

Now in projected SGDM, we have
\begin{align}
\begin{split}
\|\wsbf_{t+1}-\tilde\wsbf_t\|^2&=\|\Pbf(\ysbf_{t+1})-\tilde\wsbf_t\|^2\nn\\&\leq\|\ysbf_{t+1}-\tilde\wsbf_t\|^2
\end{split}
\end{align} since projection a point onto $\Omega$ moves it closer to any point in $\Omega$. This shows inequality \eqref{recb} holds, and the convergence results do not change.\epr

\cref{thm:sconopt}  bounds the optimization error, \ie the expected difference between the empirical risk achieved by~\ref{update} and the global minimum. Upon setting $\mu=0$ and $\gamma=0$ in~\cref{conv}, we can recover the classical bound on optimization error for SGD~\citep{nemirovski}, \citep[Theorem 5.2]{Hardt}. The first two terms in~\cref{conv} vanish as $T$ increases. The terms with negative sign improve the convergence due to the strongly convexity. The last term depends on the step-size, $\alpha$, the momentum parameter $\mu$, and the Lipschitz constant $L$. This term can be reduced by selecting $\alpha$ sufficiently small.

\section{Upper bound on true risk}
\label{app:prop:optlearning}
We now study how the uniform stability results in an upper bound on the true risk in the strongly convex case. We also compare the final results with SGD with no momentum and we show that one can achieve tighter bounds by using~\ref{update} than vanilla SGD.

The expected true risk estimate under parameter $\hat\wbf_T$ can be decomposed into a stability error term and an optimization one. In~\cref{app:thm:sconopt}, we present an upper bound on the optimization error for strongly convex loss. The optimization error reflects the optimality gap when we optimize the empirical risk under some step-size and momentum. By combining the result~\cref{app:thm:sconopt} and our stability error bound, and adjusting the hyper parameters, we minimize the upper bound on the expected true risk estimate.

In the following lemma, we show that stability results similar to~\cref{thm:sconstab} hold even if we consider the average parameter $\hat\wbf_T$ instead of $\wbf_T$. In other words, the same upper bound holds even if $\hat\wbf_T$ is considered as the output of algorithm $A$.

\blm\label[lemma]{thm:savg}
Suppose that that  $\loss$ satisfies~\cref{assu:LipSm,assu:SC} and~\ref{projupdate} is executed for $T$ steps with step-size $\alpha$ and momentum $\mu$. Provided that $\frac{\alpha\beta\gamma}{\beta+\gamma}-\frac{1}{2}\leq\mu< \frac{\alpha\beta\gamma}{3(\beta+\gamma)}$ and $\alpha\leq\frac{2}{\beta+\gamma}$, then the average of the first $T$ steps of~\ref{projupdate} satisfies $\epsilon_s$-uniform stability with~\cref{epss}.
\elm
\bpr
Let us define $\hat\wsbf_t=\frac{1}{t}\sum_{k=1}^t\wsbf_k$ and $\hat\delta_t\defeq\|\hat\wsbf_t-\hat\wsbf_t'\|$ where $\hat\wsbf_t'$ is obtained as specified in the proof of~\cref{thm:sconstab}. Following the proof of~\cref{thm:sconstab}, we have
\begin{align}\label{deltatildesc}
\E[\tilde\delta_{k+1}]\leq\Big(1+3\mu-\frac{\alpha\beta\gamma}{\beta+\gamma}\Big)\E[\tilde\delta_k]+\frac{2\alpha L}{n}.
\end{align} for $k=0,\ldots,T$. Defining $\overline\delta_t\defeq \frac{\sum_{k=1}^t\tilde\delta_k}{t}$, we have $\hat\delta_T\leq\overline\delta_T$ by the triangle inequality. Summing~\cref{deltatildesc} for $k=0,\ldots,T$ and dividing by $T$, we have $\E[\hat\delta_T]\leq\E[\overline\delta_T]\leq\frac{2\alpha L(\beta+\gamma)}{n\big(\alpha\beta\gamma-3\mu(\beta+\gamma)\big)}$. Applying the $L$-Lipschitz property on $\loss(\cdot,\zbf)$, we have
\begin{align}
\E[|\loss(\hat\wsbf_T;\zbf)-\loss(\hat\wsbf'_T;\zbf)|]\leq \frac{2\alpha L^2(\beta+\gamma)}{n\big(\alpha\beta\gamma-3\mu(\beta+\gamma)\big)},
\end{align} which holds for all $\Sc$, $\Sc'$, and $\zbf$.
\epr

Adding the stability error following~\cref{thm:savg}, we have
\begin{align}\label{opterr}
\begin{split}
\E_{\Sc,A}[R(\hat\wbf_T)]\leq \E_{\Sc,A}[R_\Sc(\hat\wbf_T)]+\epsilon_s\leq\E_{\Sc,A}[R_\Sc(\wbf_\Sc^*)]+\epsilon_{\opt}+\epsilon_s
\end{split} 
\end{align} where $\epsilon_{\opt}\defeq \E_{\Sc,A}[R_\Sc(\hat\wbf_T)-R_\Sc(\wbf_\Sc^*)].$

Note that there is a tradeoff between the optimization error and stability one. We can balance these errors to achieve reasonable expected true risk.

\bth\label{prop:optlearning}
Suppose  that  $\loss$ satisfies~\cref{assu:LipSm,assu:SC} and  that~\ref{projupdate} is executed   for $T$ steps with constant step-size $\alpha=C/T^q$ for $q\in [\frac{1}{2},1)$ and momentum $\mu$, satisfying the conditions in~\cref{thm:sconstab} with $\mu = o(\alpha\gamma)$. Then, the risk $\E[R(\hat\wsbf_T)]-\E[R_\Sc(\wsbf_\Sc^*)]$ goes to zero  as $T$ and $n$ increase 
with the rate:
\begin{align}\label{excess}\tag{Excess Risk}
\E[R(\hat\wsbf_T)]-\E[R_\Sc(\wsbf_\Sc^*)]= \Oc\Big(\max\Big\{\frac{1}{T^{q\ind\{q\leq 1/2\}+(1-q)\ind\{q> 1/2\}}
},\frac{1}{n}\Big\}\Big).   
\end{align}
\eth
\bpr By our convergence analysis in~\cref{thm:sconopt}, we have
\begin{align}\nn
\epsilon_{\opt}\leq\frac{\mu W_0}{(1-\mu)T}+\frac{(1-\mu)W_1}{2\alpha T}-\frac{\gamma W_2}{2}-\frac{\mu\gamma W_3}{2(1-\mu)}+\frac{\alpha L^2}{2(1-\mu)}.
\end{align}

By our stability analysis in~\cref{thm:savg}, we have
\begin{align}\nn
\epsilon_s\leq \frac{2\alpha L^2(\beta+\gamma)}{n\big(\alpha\beta\gamma-3\mu(\beta+\gamma)\big)}.
\end{align}
 Adding the upper bounds of $\epsilon_{\opt}$ and $\epsilon_s$ above, we have
\begin{align}\label{tunedalphaeq}
\begin{split}
\E_{\Sc,A}[R(\hat\wsbf_T)]&\leq \E_{\Sc,A}[R_\Sc(\wsbf_\Sc^*)]+\frac{\mu W_0}{(1-\mu)T}+\frac{(1-\mu)W_1}{2\alpha T}-\frac{\gamma W_2}{2}\\
&\quad-\frac{\mu\gamma W_3}{2(1-\mu)}+\frac{\alpha L^2}{2(1-\mu)}+\frac{2\alpha L^2(\beta+\gamma)}{n\big(\alpha\beta\gamma-3\mu(\beta+\gamma)\big)}.
\end{split}
\end{align}

We note that the condition $\mu <\frac{\alpha\beta\gamma}{3(\beta+\gamma)}$ in~\cref{thm:sconstab} implies that $\mu<\alpha\gamma/3$. For sufficiently small $\mu = o(\alpha\gamma)$, the last term in the upper bound becomes independent of $\alpha$ and we have 
\begin{align}\nn
\frac{2\alpha L^2(\beta+\gamma)}{n\big(\alpha\beta\gamma-3\mu(\beta+\gamma)\big)}=\Oc(1/n).    
\end{align}   

Then for any $\alpha=C/T^q$ for $q\in [\frac{1}{2},1)$, $T$ and $n$, the upper bound on the risk goes to zero as $T$ and $n$ increase with the rate in~\cref{excess}.
\epr

\cref{prop:optlearning} provides a bound on the  expected true risk of~\ref{projupdate} in terms of the global minimum of the empirical risk. %

\section{Generalization error of~\ref{earlyupdate} with  $\alpha_t=\alpha_0/\sqrt{t}$}\label{app:sqrt}

We  establish an upper bound on the generalization error of~\ref{earlyupdate} with the larger step size $\alpha_t=\alpha_0/\sqrt{t}$, which is a common choice in the optimization literature~\citep{bubeck2015convex}.

\bth\label{thm:sqrtlrstab}
Suppose {that  $\loss$ satisfies~\cref{assu:LipSm} and} that  the \ref{earlyupdate} update is executed for $T$ steps with step-size $\alpha_t=\alpha_0/\sqrt{t}$ and some constant {$\mu_d\in(0,1]$} in the first $t_d$ steps. Then, for any $1 \leq \tilde t \leq t_d \leq T$,~\ref{earlyupdate} satisfies $\epsilon_s$-uniform stability with
\begin{align}\label{sqrtlrstab}
\!\!\!\!\!\!\epsilon_s\leq\frac{2\alpha_0L^2\sqrt{\pi}}{n\sqrt{2\mu_d}}\exp(u\sqrt{T})\breve h(\mu_d,t_d)+\frac{\tilde t M}{n}+\frac{2L^2}{\beta(n-1)}\exp\Big(u\big(\sqrt{T}-\sqrt{\tilde t}\big)\Big)
\end{align}
where $\breve h(\mu_d,t_d)=\exp(2\mu_dt_d+u^2/(8\mu_d))\big(\Phi(\sqrt{2\mu_d}(\sqrt{t_d}+\frac{u}{4\mu_d}))-\Phi(\sqrt{2\mu_d}(\sqrt{\tilde t}+\frac{u}{4\mu_d}))\big)$,  $\Phi(x)=\erf(x):=\frac{2}{\sqrt{\pi}}\int_{0}^{x}\exp(-t^2)\ud t$,  $u=(1-\frac{1}{n})\alpha_0\beta$, and $M=\sup_{\wbf,\zbf}\loss(\wbf;\zbf)$. 
\eth

\bpr Similar to the proof of~\cref{thm:earlymnconstab}, we have the following inequality:
\begin{align}\nn
\tilde\Delta_{t+1,\tilde t}\leq\big(1+2\mu_t+(1-1/n)\alpha_t\beta\big)\tilde\Delta_{t,\tilde t}+\frac{2\alpha_t L}{n}.
\end{align}
 Noting that $\tilde\Delta_{t,\tilde t}\geq \Delta_{t,\tilde t}$ for all $t\geq \tilde t$, we have $\E[\Delta_{T,\tilde t}]\leq S_3+S_4$ where
\begin{align}\nn
S_{3}=\sum_{t=\tilde t+1}^{t_d}\prod_{p=t+1}^{T}\Big(1+2\mu_p+\big(1-\frac{1}{n}\big)\frac{\alpha_0\beta}{\sqrt{p}}\Big)\frac{2\alpha_0 L}{n\sqrt{t}}
\end{align} and
\begin{align}\nn
S_{4}=\sum_{t=t_d+1}^{T}\prod_{p=t+1}^{T}\Big(1+2\mu_p+\big(1-\frac{1}{n}\big)\frac{\alpha_0\beta}{\sqrt{p}}\Big)\frac{2\alpha_0 L}{n\sqrt{t}}.
\end{align}
Substituting $\mu_p=\mu_d$ for $p=1,\ldots,t_d$, we can find an upper bound on $S_3$ as follows:
\begin{align}%
S_{3}&=\sum_{t=\tilde t+1}^{t_d}\prod_{p=t+1}^{T}\Big(1+2\mu_p+\big(1-\frac{1}{n}\big)\frac{\alpha_0\beta}{\sqrt{p}}\Big)\frac{2\alpha_0 L}{n\sqrt{t}}\nn\\
&\leq \sum_{t=\tilde t+1}^{t_d}\prod_{p=t+1}^{T}\exp\Big(2\mu_p+\big(1-\frac{1}{n}\big)\frac{\alpha_0\beta}{\sqrt{p}}\Big)\frac{2\alpha_0 L}{n\sqrt{t}}\nn\\
&\leq \sum_{t=\tilde t+1}^{t_d}\exp\Big(2\mu_d(t_d-t)+\big(1-\frac{1}{n}\big)\alpha_0\beta\big(\sqrt{T}-\sqrt{t}\big)\Big)\frac{2\alpha_0 L}{n\sqrt{t}}\nn\\
&\leq \frac{2\alpha_0L}{n}\exp(u\sqrt{T}+2\mu_dt_d)\int_{\tilde t}^{t_d}\frac{\exp(-2\mu_d t-u\sqrt{t})}{\sqrt{t}}\ud t\nn\\
&\leq \frac{2\alpha_0L}{n}\exp(u\sqrt{T}+2\mu_dt_d+u^2/(8\mu_d))\int_{\tilde t}^{t_d}\frac{\exp\big(-2\mu_d(\sqrt{t}+u/(4\mu_d))^2\big)}{\sqrt{t}}\ud t\nn\\
&=\frac{2\alpha_0L\sqrt{\pi}}{n\sqrt{2\mu_d}}\exp(u\sqrt{T}+2\mu_dt_d+u^2/(8\mu_d))\big(\Phi(\sqrt{2\mu_d}(\sqrt{t_d}+\frac{u}{4\mu_d}))-\Phi(\sqrt{2\mu_d}(\sqrt{\tilde t}+\frac{u}{4\mu_d}))\big)\nn
\end{align} where the last line follows~\citep[Eq. 3.321]{IntTable}.

We can also find an  upper bound on $S_4$ as follows:
\begin{align}\label{S4UBsqrtlr}
\begin{split}
S_{4}&=\sum_{t=t_d+1}^{T}\prod_{p=t+1}^{T}\Big(1+\big(1-\frac{1}{n}\big)\frac{\alpha_0\beta}{\sqrt{p}}\Big)\frac{2\alpha_0 L}{n\sqrt{t}}\\
&\leq \frac{2L}{\beta(n-1)}\exp\Big(u\big(\sqrt{T}-\sqrt{t_d}\big)\Big)\\
&\leq \frac{2L}{\beta(n-1)}\exp\Big(u\big(\sqrt{T}-\sqrt{\tilde t}\big)\Big).
\end{split}
\end{align}

Replacing $\Delta_{T,\tilde t}$ with its upper bound in~\cref{nonconine}, we obtain~\cref{sqrtlrstab}.

By its definition, we have $\Phi(x)\leq 1$. We also note that $1-\exp(-x^2)\leq \Phi(x)$ for $x>0$ following the upper bound developed for $1-\erf$ in~\citep{chiani2003new}. Applying both lower bound and upper bound on $\Phi$ in~\cref{sqrtlrstab} and after rearranging the terms, we have
\begin{align}\label{sqrtlrstabsim}
\epsilon_s\leq\Big(\frac{2\alpha_0L^2\sqrt{\pi}}{n\sqrt{2\mu_d}}\exp\big(2\mu_d(t_d-\tilde t)\big)
+\frac{2L^2}{\beta(n-1)}
\Big)\exp\Big(u\big(\sqrt{T}-\sqrt{\tilde t}\big)\Big)+\frac{\tilde t M}{n}.
\end{align}
\epr

\bcr\label{appcr:sqrtlr} Suppose, in~\cref{thm:sqrtlrstab}, we set $t_d=\tilde t^*+K$for some constant $K$ where $\tilde t^*$ satisfies:
\begin{align}\label{opt_tilde_sqrtlr}
M\exp(u\sqrt{\tilde t^*})\sqrt{\tilde t^*}=\Big(\frac{u\alpha_0L^2\sqrt{\pi}}{\sqrt{2\mu_d}}+L^2\alpha_0\Big)\exp(u\sqrt{T}).\end{align} Then the generalization error of~\ref{earlyupdate} for $T$ steps with  $\alpha_t=\alpha_0/\sqrt{t}$ is upper bounded by $\Oc\Big(\frac{\exp(u\sqrt{T}/(u+1)+\mu_d)}{n}\Big)$. 
\ecr
\bpr  Note that we can minimize: \begin{align}\nn
\min_{1\leq \tilde t\leq t_d}\frac{\tilde t M}{n}+\Big(\frac{2\alpha_0L^2\sqrt{\pi}}{n\sqrt{2\mu_d}}\exp\big(2\mu_dK\big)
+\frac{2L^2}{\beta(n-1)}
\Big)\exp\Big(u\big(\sqrt{T}-\sqrt{\tilde t}\big)\Big)    
\end{align}  
by optimizing $\tilde t$ after setting $t_d=\tilde t+K$ where the objective is the upper bound in~\cref{sqrtlrstab}. We note that an optimal $\tilde t^*$ satisfies~\cref{opt_tilde_sqrtlr}, which does not have an analytic solution but can be solve numerically. Instead, we consider a suboptimal solution by taking $\ln$ from both sides of~\cref{opt_tilde_sqrtlr} and applying the well-known inequality $\ln(x+1)\leq x$, $\forall x\geq -1$, which leads to:

\begin{align}\label{subopt_tilde_sqrtlr}
\sqrt{\tilde t} = \frac{\ln\Big(\big(\frac{u\alpha_0L^2\sqrt{\pi}}{\sqrt{2\mu_d}}+L^2\alpha_0\big)/M\Big)}{u+1}+\frac{u\sqrt{T}}{u+1}.
\end{align}
 
Substituting~\cref{subopt_tilde_sqrtlr} into~\cref{sqrtlrstab} completes the proof.
\epr 

\section{Additional experiments}\label{app:exp}
\begin{figure}[t]
\centering
\includegraphics[width=0.7\textwidth]{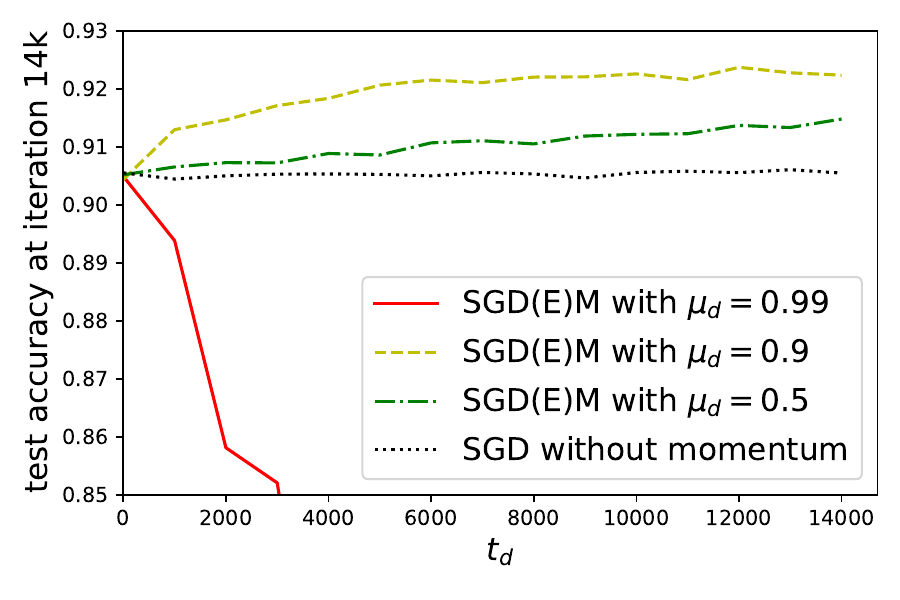}
\caption{Test accuracy of a feedforward fully connected neural network for notMNIST dataset.}
\label{NF5b}
\end{figure}

In~\cref{NF5b}, we plot the test accuracy versus $t_d$ of~\ref{earlyupdate} and~\ref{update} (which is a special case of~\ref{earlyupdate} with $t_d=T$) for the notMNIST dataset for different $\mu_d$ values. We observe dramatic decrease in the test accuracy for $\mu_d=0.99$, which is consistent with our convergence analysis in \cref{thm:earlymnconconv}.

\begin{figure}[t]
\centering
\includegraphics[width=0.7\textwidth]{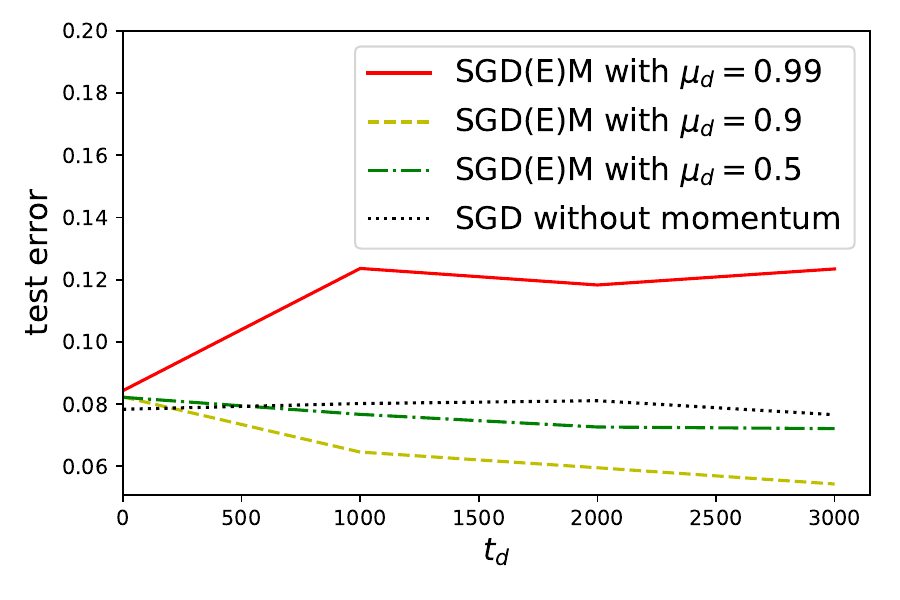}
\caption{Test error of logistic regression for notMNIST dataset.}
\label{NF7a}
\end{figure}
\begin{figure}[t]
\centering
\includegraphics[width=0.7\textwidth]{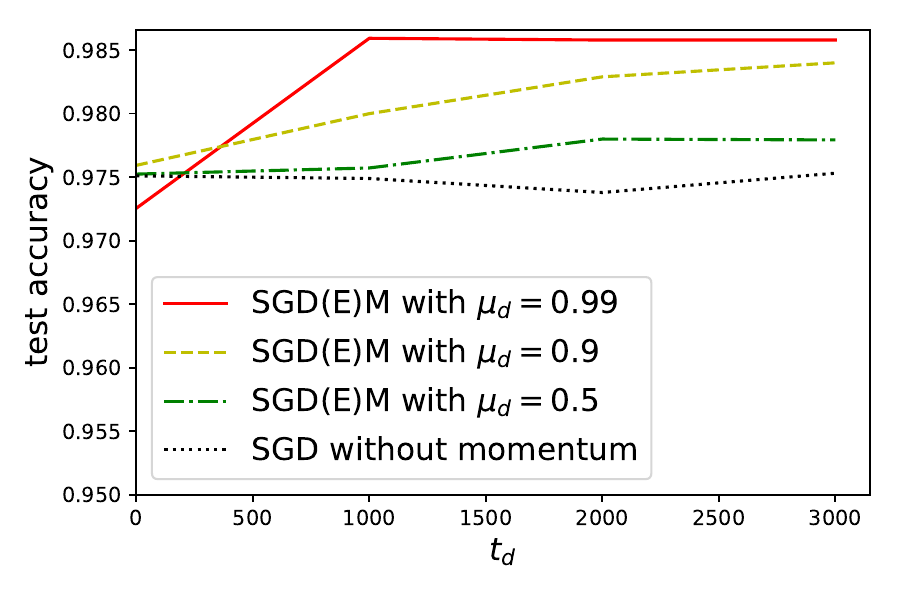}
\caption{Test accuracy of logistic regression for notMNIST dataset.}
\label{NF7b}
\end{figure}

\begin{figure}[t]
\centering
\includegraphics[width=0.7\textwidth]{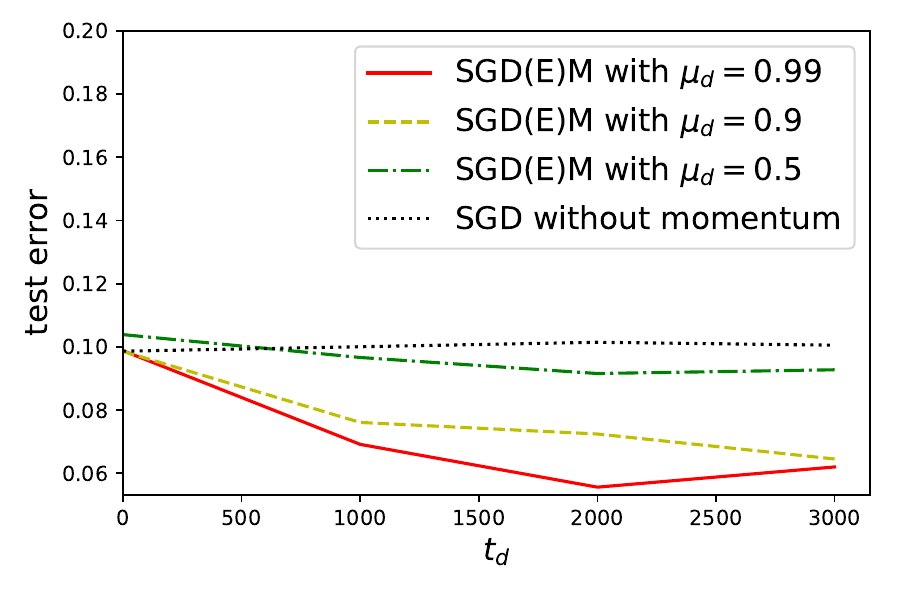}
\caption{Test error of logistic regression for MNIST dataset.}
\label{NF8a}
\end{figure}
\begin{figure}[t]
\centering
\includegraphics[width=0.7\textwidth]{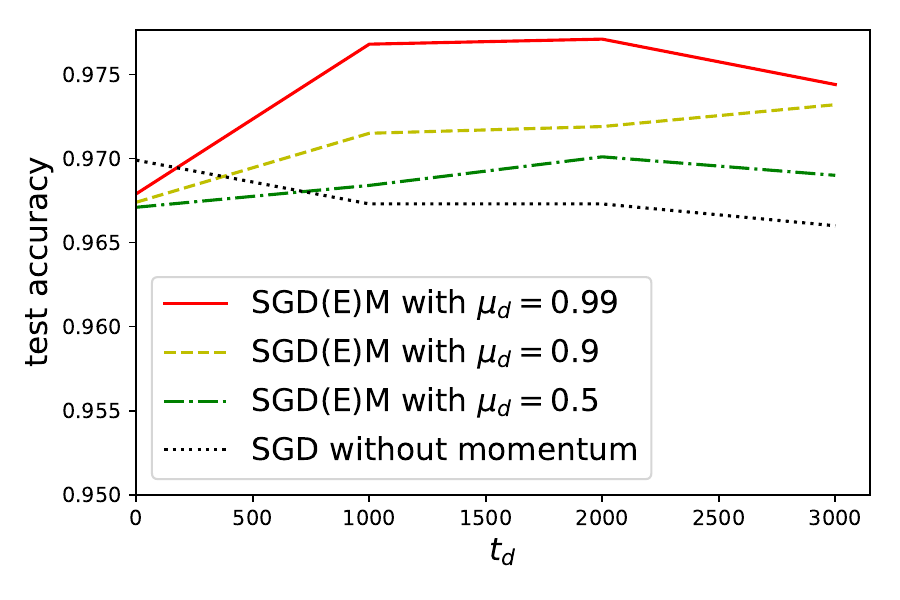}
\caption{Test accuracy of logistic regression for MNIST dataset.}
\label{NF8b}
\end{figure}

We now study the performance of~\ref{earlyupdate} for a smooth and strongly convex loss function. We train a logistic regression model with the weight decay regularization using~\ref{earlyupdate} for binary classification on the two-class notMNIST and MNIST datasets that contain the images from letter classes ``C'' and ``J'', and digit classes ``2'' and ``9'', respectively. We set the step-size $\alpha=0.01$. The weight decay coefficient and the minibatch size are set to 0.001 and 10, respectively. We use 100~\ref{earlyupdate} realizations to evaluate the average performance.

We plot the test error and test accuracy versus $t_d$ under~\ref{earlyupdate} for the notMNIST dataset in~\cref{NF7a,NF7b}, respectively. We show the same performance measures for the MNIST dataset in~\cref{NF8a,NF8b} respectively. We observe that, unlike the case of nonconvex loss functions, it does not hurt to
add momentum for the entire training. In the following, we focus on~\ref{update} with the classical momentum update rule for a smooth and strongly convex loss function for the notMNIST dataset.

\begin{figure}[t]
\centering
\includegraphics[width=0.7\textwidth]{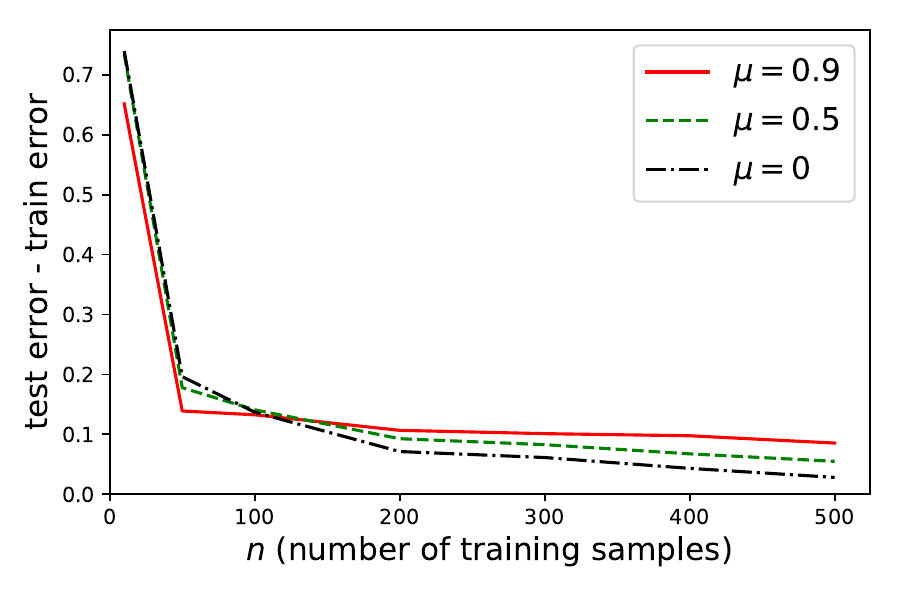}
\caption{Generalization error (cross entropy) of logistic regression for notMNIST dataset with $T=1000$ iterations.}
\label{NF1}
\end{figure}
\begin{figure}[t]
\centering
\includegraphics[width=0.7\textwidth]{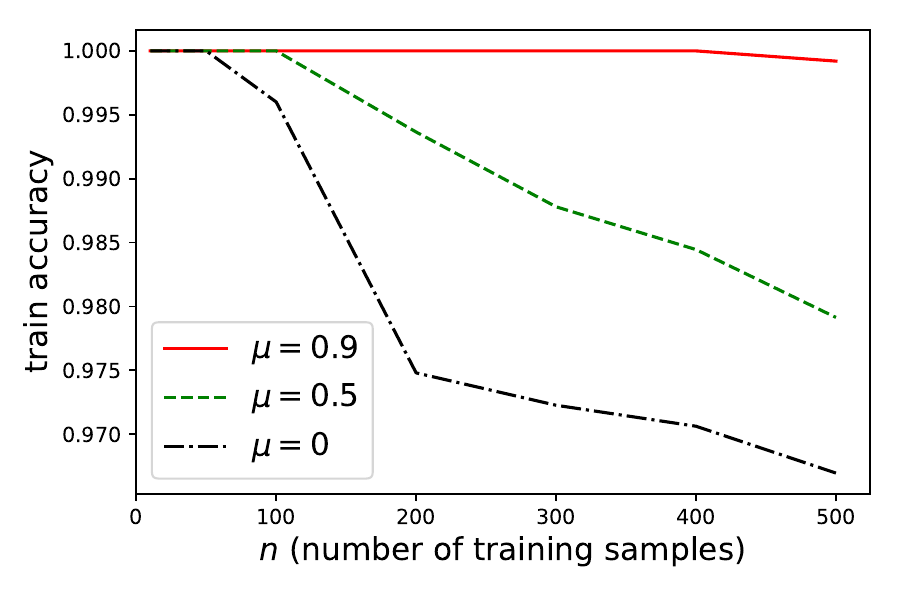}
\caption{Training accuracy of logistic regression for notMNIST dataset with $T=1000$ iterations.}
\label{NF2a}
\end{figure}

In the following, we focus on~\ref{update} with the classical momentum update rule for a smooth and strongly convex loss function on notMINIST.

We compare the training and generalization performance of SGD without momentum with that of~\ref{update} under $\mu=0.5$ and $\mu=0.9$, which are common momentum values used in practice~\citep[Section 8.3.2]{DLbook}. %

We show in~\cref{NF1} the generalization error (w.r.t. cross entropy)  versus the number of training samples, $n$, under~\ref{update} with fixed $T=1000$ iterations for $\mu=0, 0.5, 0.9$. In~\cref{NF2a}, we plot the training accuracy as a function of the number of training samples for the same dataset.  First, we observe that the generalization error  decreases as $n$ increases  for all values of $\mu$, which is also suggested by our stability upper bound in~\cref{thm:sconstab}.  In addition, for sufficiently large $n$, we observe that the generalization error  increases with $\mu$, consistent with~\cref{thm:sconstab}.
The training accuracy also improves by adding momentum as illustrated in~\cref{NF2a}.

\begin{figure}[t]
\centering
\includegraphics[width=0.7\textwidth]{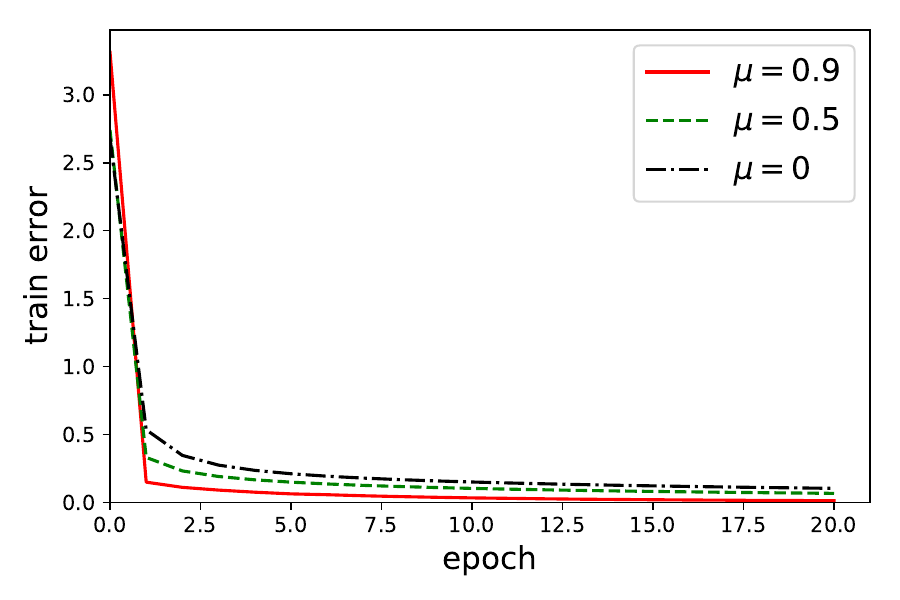}
\caption{Training error (cross entropy) of logistic regression  for notMNIST dataset with $n=500$.}
\label{NF3}
\end{figure}
\begin{figure}[t]
\centering
\includegraphics[width=0.7\textwidth]{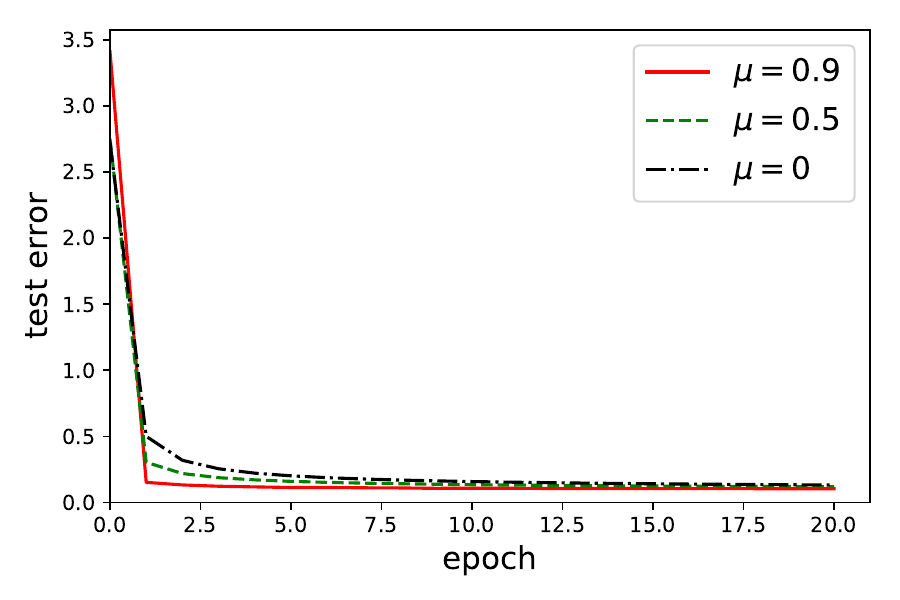}
\caption{Test error (cross entropy) of logistic regression  for notMNIST dataset with $n=500$.}
\label{NF4}
\end{figure}

\begin{figure}[t]
\centering
\includegraphics[width=0.7\textwidth]{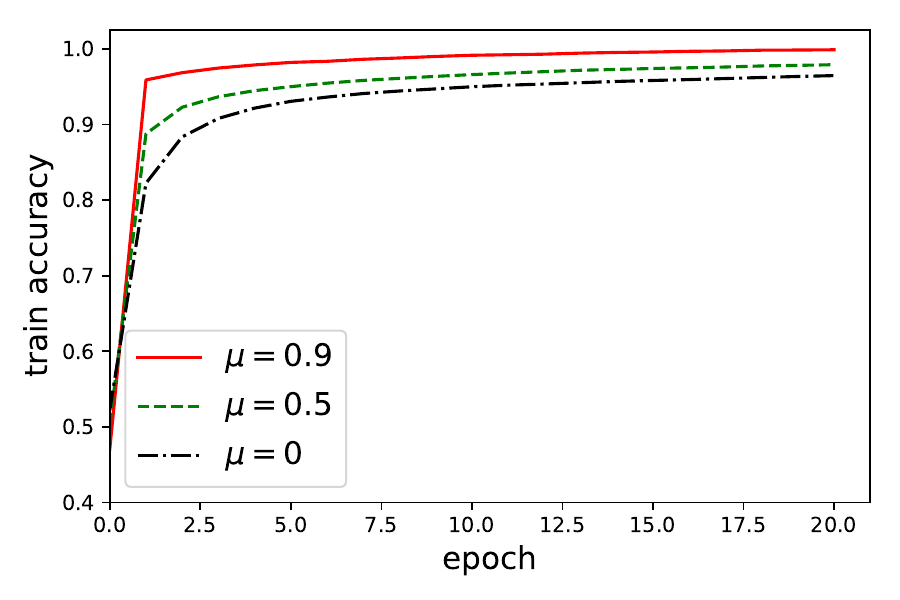}
\caption{Training accuracy of logistic regression  for notMNIST dataset with $n=500$.}
\label{NF3a}
\end{figure}

In order to study the optimization error of~\ref{update}, we show in~\cref{NF3,NF4}, the training error and test error, respectively, versus the number of epochs, under~\ref{update} trained with $n=500$ samples. We plot the classification accuracy for training dataset in~\cref{NF3a}. We observe that the training error decreases as the number of epochs increases for all values of $\mu$, which is consistent with the convergence analysis in~\cref{thm:sconopt}. Furthermore, as expected, we see that adding momentum improves the training error and accuracy. However, as the number of epochs increases, we note that  the benefit of  momentum on the test error becomes negligible. This happens because  adding momentum also results in a  higher generalization error thus offsetting the gain in training error.

\vskip 0.2in
\bibliography{Ref}

\end{document}